\documentclass[10pt,twocolumn,letterpaper]{article}

\usepackage{cvpr}              

\usepackage{graphicx}
\usepackage{amsmath}
\usepackage{amssymb}
\usepackage{booktabs}

\usepackage[utf8]{inputenc} 
\usepackage[T1]{fontenc}    
\usepackage{url}            
\usepackage{amsfonts}       
\usepackage{nicefrac}       
\usepackage{microtype}      

\usepackage{balance}
\usepackage{multirow}
\usepackage{color}
\usepackage{comment}
\usepackage{xcolor}
\usepackage{helvet}
\usepackage{courier}
\usepackage{balance}
\usepackage{amsmath,amsthm,amssymb}
\usepackage{textcomp,booktabs}
\usepackage{footmisc}
\usepackage[normalem]{ulem}
\usepackage{multirow}
\usepackage{setspace}
\usepackage{bigstrut}
\usepackage{array}
\usepackage{bm}
\usepackage{bbm}
\usepackage{url}
\usepackage{threeparttable}
\usepackage{algorithm}
\usepackage{algorithmic}
\usepackage{wrapfig}

\newcommand{\rb}{\boldsymbol{r}}

\newcommand{\z}{\boldsymbol{z}}

\newcommand{\Rb}{\mathbf{R}}

\usepackage{colortbl}
\definecolor{myy}{RGB}{126,95,0}
\definecolor{mygray}{gray}{.9}
\definecolor{bblue}{RGB}{30,80,120}
\definecolor{mygray1}{gray}{.7}

\newtheorem{principle}{Principle}

\newcommand\blfootnote[1]{%
  \begingroup
  \renewcommand\thefootnote{}\footnote{#1}%
  \addtocounter{footnote}{-1}%
  \endgroup
}

\newcommand{\lv}[1]{{\color{black}{#1}}}
\usepackage[pagebackref,breaklinks,colorlinks]{hyperref}

\usepackage[capitalize]{cleveref}
\crefname{section}{Sec.}{Secs.}
\Crefname{section}{Section}{Sections}
\Crefname{table}{Table}{Tables}
\crefname{table}{Tab.}{Tabs.}

\usepackage{fdsymbol}
\def\cvprPaperID{4653} 
\def\confName{CVPR}
\def\confYear{2022}

\begin{document}

\title{Causality Inspired Representation Learning for Domain Generalization}


\author{
\textbf{Fangrui Lv}\textsuperscript{\rm 1}
~
\textbf{Jian Liang}\textsuperscript{\rm 2}
~
\textbf{Shuang Li}\textsuperscript{\rm 1,$*$} 
~
\textbf{Bin Zang}\textsuperscript{\rm 1}
~
\textbf{Chi Harold Liu}\textsuperscript{\rm 1}
~
\textbf{Ziteng Wang}\textsuperscript{\rm 3}
~
\textbf{Di Liu}\textsuperscript{\rm 2}
\\ [0.25cm]
\textsuperscript{\rm 1} Beijing Institute of Technology, China
\ \ 
\textsuperscript{\rm 2}Alibaba Group, China
\ \ 
\textsuperscript{\rm 3}Yizhun Medical AI Co., Ltd, China
\\[0.1cm]
{$^1$ \tt\small \{fangruilv,shuangli,binzang\}@bit.edu.cn, liuchi02@gmail.com}\\
{$^2$ \tt\small \{xuelang.lj,wendi.ld\}@alibaba-inc.com}
~~
{$^3$ \tt\small ziteng.wang@yizhun-ai.com}
}

\maketitle

\begin{abstract}
Domain generalization (DG) is essentially an out-of-distribution problem, aiming to generalize the knowledge learned from multiple source domains to an unseen target domain. The mainstream is to leverage statistical models to model the dependence between data and labels, intending to learn representations independent of domain.
Nevertheless, the statistical models are superficial descriptions of reality since they are only required to model dependence instead of the intrinsic causal mechanism. When the dependence changes with the target distribution, the statistic models may fail to generalize.
In this regard, we introduce a general structural causal model to formalize the DG problem. Specifically, we assume that each input is constructed from
a mix of causal factors (whose relationship with the label is invariant across domains) and non-causal factors (category-independent), and only the former cause the classification judgments. Our goal is to extract the causal factors from inputs and then reconstruct the invariant causal mechanisms. 
However, the theoretical idea is far from practical of DG since the required causal/non-causal factors are unobserved. We highlight that ideal causal factors should meet three basic properties: separated from the non-causal ones, jointly independent, and causally sufficient for the classification. 
Based on that, we propose a Causality Inspired Representation Learning (CIRL) algorithm that enforces the representations to satisfy the above properties and then uses them to simulate the causal factors, which yields improved generalization ability. 
Extensive experimental results on several widely used datasets verify the effectiveness of our approach.
\blfootnote{$*$ Corresponding author.}
\footnote{\quad Code is available at "https://github.com/BIT-DA/CIRL".}
\vspace{-4mm}
\end{abstract}

\section{Introduction}
\label{sec:intro}

In recent years, with the increasing complexity of tasks in real world, out-of-distribution (OOD) problem has raised a severe challenge for deep neural networks based on the i.i.d. hypothesis~\cite{BCDM,ParetoDA,GDCAN}. Directly applying the model trained on source domain to an unseen target domain with different distribution typically suffers from a 
catastrophic performance degradation \cite{RTN,MMAN,imagenet-c,robustness}.
In order to deal with the domain shift problem, Domain Generalization (DG) has attracted increasing attention, which aims to generalize the knowledge extracted from multiple source domains to an unseen target domain \cite{DICA,MMD-AAE,MetaReg,DBA}.

In order to improve generalization 
\lv{capability}, many DG methods have been proposed, which can be roughly categorized into invariant representation learning \cite{MTA,MMD-AAE,CIAN,CCSA}, domain augmentation \cite{CrossGrad,AdvAug, MixStyle, FACT}, meta-learning \cite{MetaReg,MASF,Ave}, etc.
Though promising results have been achieved, there exists one intrinsic problem with them. 
These efforts merely try to make up for the problems caused by 
\lv{OOD} data and model the statistical dependence between data and labels without explaining the underlying causal mechanisms. It has been argued recently \cite{2017Elements} that such practices may not be sufficient, and generalizing well outside the i.i.d. setting requires learning not mere statistical dependence between variables, but an underlying causal model \cite{CRLS,LICM,2017Elements,causality,anti-causal,2000Causation}. 
For instance, in an image classification task, it is very likely that all the giraffes are on the grass,
showing high statistical dependence, 
which could easily mislead the model to make wrong predictions when the background 
\lv{varies} in target domain.
After all, the characteristics of giraffes such as head, neck, etc., 
instead of the background make a giraffe \textit{giraffe}.

\begin{wrapfigure}{r}{0.23\textwidth}
  \vspace{-6mm}
    \begin{center}
      \includegraphics[width=0.23\textwidth]{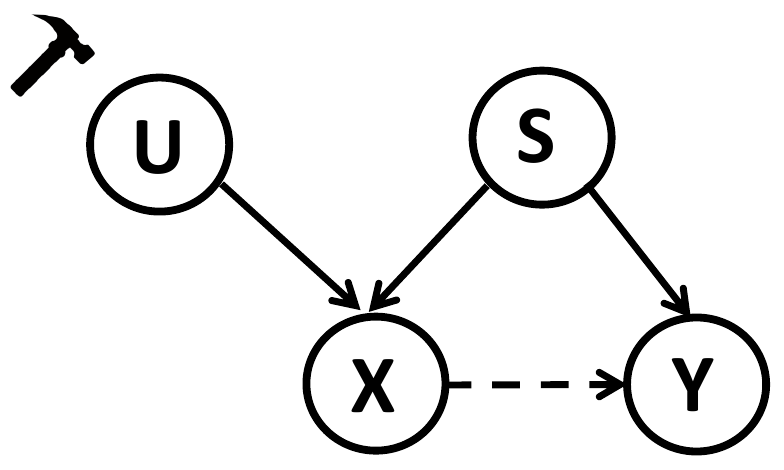}
    \end{center}
    \vspace{-6mm}
    \caption{SCM of DG. The \textbf{solid arrow} indicates that the parent node causes the child one; while the \textbf{dash arrow} means there exists statistical dependence.}
    \label{fig:scm}
  \vspace{-6mm}
  \end{wrapfigure}
In this paper, we introduce a structural causal model (SCM) \cite{scm} to formalize the DG problem,
aiming to excavate the intrinsic causal mechanisms between data and labels, and achieve better generalization ability. 
Specifically, we assume the category-related information in data as causal factors, 
whose relationship with the label is independent of domain,
e.g., \textit{"shape"} in digit recognition.
While the information independent of category is assumed as non-causal factors,
which is generally domain-related information, e.g., \textit{"handwriting style"} in digit recognition. Each raw data $X$ is constructed from a mix of causal factors $S$ and non-causal factors $U$ 
and only the former causally effects the category label $Y$, as shown in Fig. \ref{fig:scm}. 
Our goal is to extract the causal factors $S$ from raw input $X$ and then reconstruct the invariant causal mechanisms, which can be done with the aid of causal intervention $P(Y|do(U),S)$. \lv{The do-operator $do(\cdot)$ \cite{2016Causal} denotes intervention upon variables.}
Unfortunately, we cannot directly factorize raw input as $X = f(S,U) $ since the causal/non-causal factors are generally unobserved and cannot be formulated, which makes the causal
inference particularly challenging \cite{B2016Causal, WSDG}.


\begin{figure}[t]
\vspace{-2mm}
  \centering
   \includegraphics[width=1.0\linewidth]{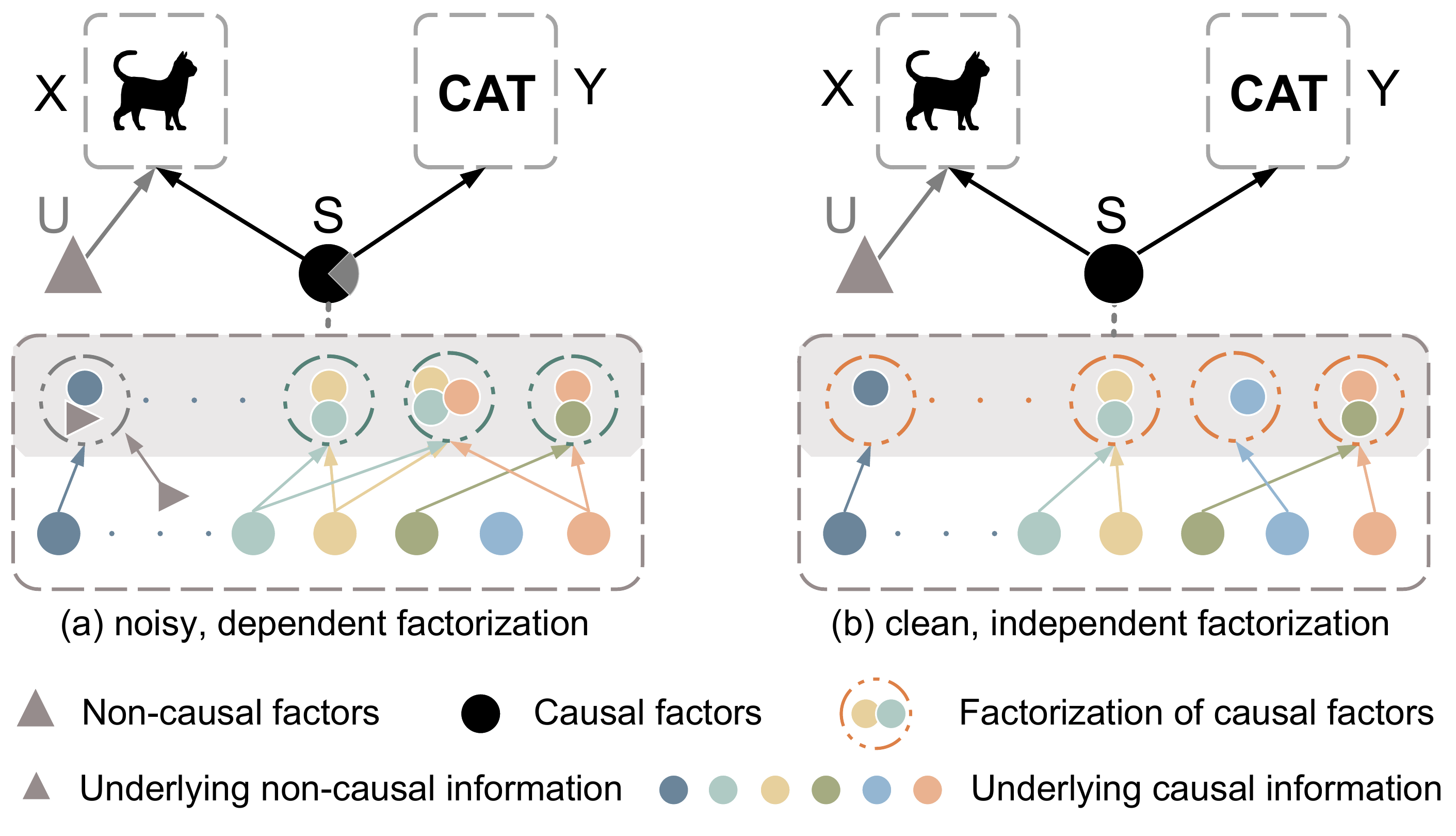}
   \vspace{-6mm}
   \caption{Illustration of the three properties of causal factors.}
   \label{fig:motivationl}
   \vspace{-6mm}
\end{figure}

In order to make the theoretical idea into practice, we  highlight that the causal factors $S$ are expected to satisfy three properties based on the researches in \cite{2017Elements,anti-causal,1956The}: 1) separated from the non-causal factors $U$; 2) the factorization of $S$ should be jointly independent; 3) causally sufficient for the classification task $X \xrightarrow{} Y$ in the sense of containing all the causal information
. As shown in Fig. \ref{fig:motivationl} (a), the mixture with $U$ causes $S$ to contain underlying non-causal information, while the jointly dependent factorization makes $S$ redundant, further leading to the miss of some underlying causal information. In contrast, the causal factors $S$ in Fig. \ref{fig:motivationl} (b) are ideal ones that meet all the requirements.
Inspired by this, we propose a Causality Inspired Representation Learning (CIRL) algorithm, enforcing the learned representations to possess the above properties and then exploiting each dimension of the representations to mimic the factorization of causal factors, which have stronger generalization ability.

Concisely, for each input, we first exploit a causal intervention module to separate the causal factors $S$ from non-causal factors $U$ via generating new data with perturbed domain-related information. The generated data have different non-causal factors $U$ but the same causal-factors $S$ compared with the original ones, so the representations are enforced to remain invariant. Besides, we propose a factorization module that makes each dimension of the representations jointly independent and then can be used to approximate the causal factors. 
Furthermore, to be causally sufficient towards classification, we design an adversarial mask module which iteratively detects dimensions that contain relatively less causal information and forces them to contain more and novel causal information via adversarial learning between a masker and the representation generator.
The contributions of our work are as follows:
\begin{itemize}
    \item We point out the insufficiency  of only modeling statistical dependence and introduce a causality-based view into DG to excavate the intrinsic causal mechanisms.  
    \item We highlight three properties that the ideal causal factors should possess, and propose a CIRL algorithm to learn causal representations that can mimic the causal factors, which have better generalization ability.
    \item Extensive experiments on several widely used datasets and analytical results demonstrate the effectiveness and superiority of our method.
\end{itemize}






\section{Related Work}
\label{sec:related}
\textbf{Domain Generalization (DG)}
aims to extract knowledge from multiple source domains that are well-generalizable to unseen target domains. 
A promising and prevalent solution is to align the distribution of domains by learning domain-invariant representation via kernel-based optimization \cite{DICA,SCA}, adversarial learning \cite{MMD-AAE,CIAN,CCSA}, second-order correlation \cite{DGCAN} or using Variational Bayes \cite{BayesDG}.
Data augmentation is also an important technique to empower the model with generalization ability by enriching source diversity. Several researches have been explored in previous works: \cite{CrossGrad} perturbs images according to adversarial gradients induced by domain discriminator. \cite{MixStyle, FACT} mix the styles of training instances across domains by mixing feature statistics \cite{MixStyle} or amplitude spectrums \cite{FACT}. \cite{L2A-OT} generates more training synthetic data by maximizing a divergence measure.
Another popular way that has been investigated is meta-learning, which simulates domain shift by dividing meta-train and meta-test domains from the original source domains \cite{MetaReg,FCN,MASF,Ave}. 
Other DG works also explore low-rank decomposition \cite{CSD}, secondary task as solving jigsaw puzzles \cite{Jigen} and gradient-guided dropout \cite{RSC}.
Different from all the methods above, we tackle DG problem from a causal viewpoint. Our method focuses on excavating intrinsic causal mechanisms by learning causal representations, which has shown better generalization ability.

\textbf{Causal Mechanism} \cite{2015Causal,2000Causality,2000Causation} focuses on the fact that statistical dependence ("seeing people take medicine suggests that he is sick") cannot reliably predict the outcome
of a counterfactual input ("stopping taking medicine does not make him healthy"). Generally, it can be viewed as components
of reasoning chains \cite{1974Causation} that provide predictions for situations that are very far from the observed distribution. In that sense, excavating causal mechanisms means acquiring robust knowledge that holds beyond the support of observed data distributions \cite{TCRL}. 
The connection between causality and generalization has gained increasing interest in the past few years \cite{Meinshausen2016Causal,2017Elements}.
Many causality-based methods have been proposed to gain invariant causal mechanisms \cite{LPMTT,Heinze-DemlM21,contrastive-ACE} or recover causal features \cite{2016Causal,Rojas-CarullaST18,ChangZYJ20,LiuH00S21} and hence improve OOD generalization. It is worth noting that they generally rely on restrictive assumptions on the causal diagram or structural equations.
Very recently, MatchDG \cite{MatchDG} introduces causality into DG literature by enforcing the inputs across domains have the same representation via contrastive learning if they are derived from the same object.
Our CIRL is related to MatchDG in its efforts to learn causal representations. However, CIRL differs in the fact that it is done explicitly with exploiting dimension-wise representations to mimic causal factors based on a much theoretical formulation and only relies on a more general causal structural model
\lv{without restrictive assumptions.}
Essentially, CIRL can be seen as causal factorization with intervention, which is clearly different from the object conditional MatchDG.

\section{Method}
\label{sec:method}
\begin{figure*}[t]
  \centering
   \includegraphics[width=0.85\linewidth]{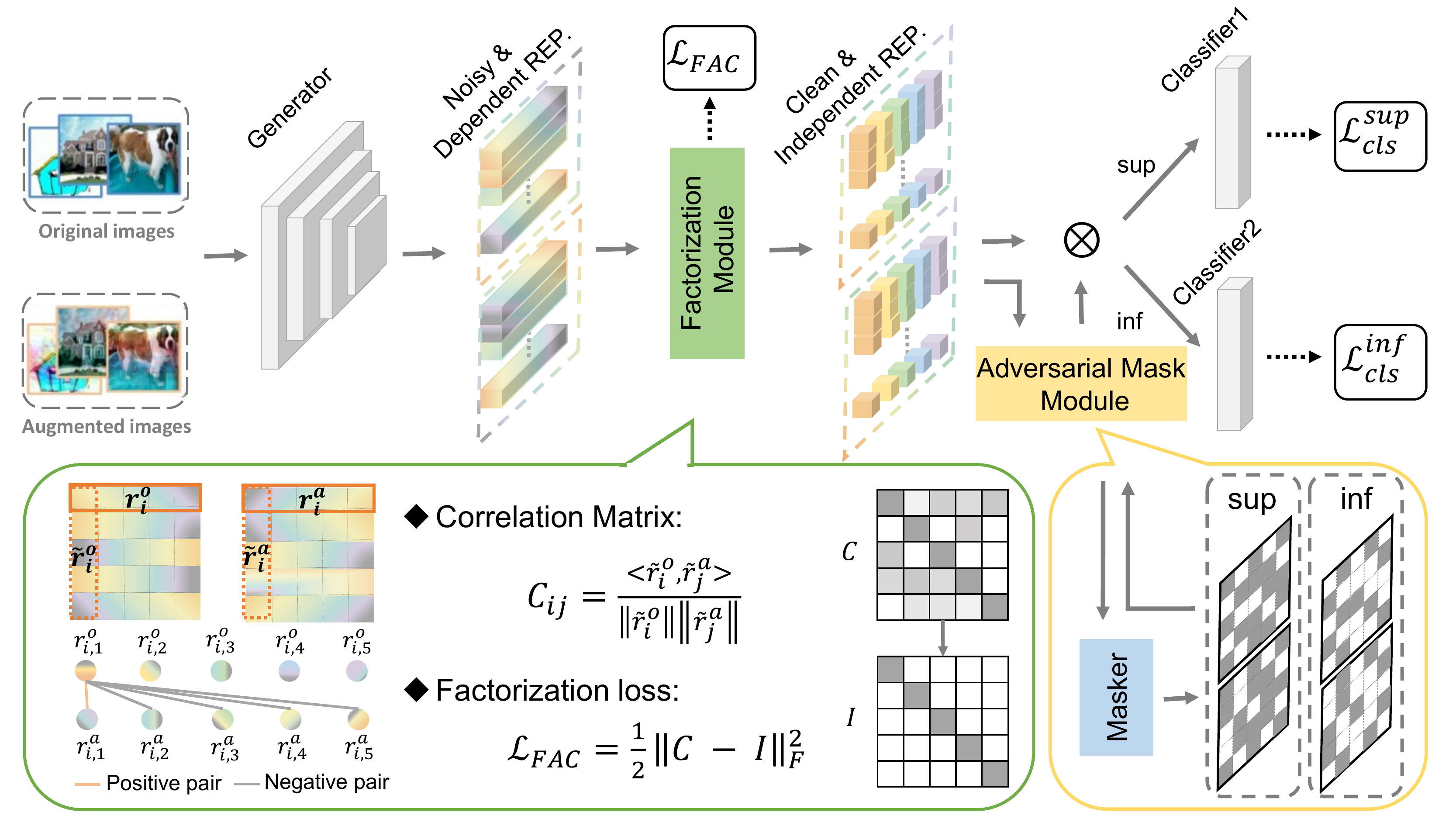}
   \vspace{-2mm}
   \caption{The framework of CIRL. We first generate augmented images by a causal intervention module with intervention upon non-causal factors. Both the representations of original and augmented images are sent to the factorization module, which imposes a factorization loss to force the representations to be separated from non-causal factors and jointly independent. At last, the adversarial mask module conducts an adversary between the generator and a masker, rendering the representations to be causally sufficient for classification. }
   \label{fig:framework}
   \vspace{-4mm}
\end{figure*}


In this section, we consider DG from the causal view with a general structural causal model as Fig.~\ref{fig:scm} shows. We demonstrate that the intrinsic causal mechanisms (formalized as conditional distributions) can be feasible to construct if the causal factors are given. However, as discussed in~\cite{Arjovsky2019}, it is hard to recover the causal factors exactly since they are unobservable. Therefore, we propose to learn causal representations based on the properties of causal factors as a mimic, while inheriting the superior generalization ability.

\subsection{DG from the Causal View}\label{subsec:sec_31}
The mainstream of DG focuses on modeling the statistical dependence between observed inputs and corresponding labels, i.e., $P(X,Y)$, which is assumed variant across domains. 
To obtain an invariant dependence, they generally enforce the distribution to be domain-invariant marginally or conditionally, i.e., minimizing the gap across domains in $P(X)$ or $P(X\mid Y)$. 
However, since the statistical dependence cannot explain the intrinsic causal mechanism between inputs and labels, it tends to vary with domain. Therefore, the learned invariant dependence among source domains may still fail on an unseen target domain.
Meanwhile, causal mechanisms usually keep stable across domains~\cite{2017Elements}.
We first articulated the connection between causality and statistical dependence as Reichenbach~\cite{1956The} claimed in Principle~\ref{CCP}.
\begin{principle}[\cite{1956The}]\label{CCP}
\textbf{Common Cause Principle}: if two observables $X$ and $Y$
are statistically dependent, then there exists a variable
$S$ that causally influences both and explains all the
dependence in the sense of making them independent
when conditioned on $S$.
\end{principle}
Based on Principle~\ref{CCP}, we formalize the following structural causal model (SCM) to describe the DG problem:
\begin{equation}\small
\begin{split}
    &X := f(S,U,V_1), S \Vbar U \Vbar V_1 , \\
    &Y := h(S ,V_2) = h(g(X),V_2), V_1 \Vbar V_2.
\end{split}
\label{assignment}
\end{equation}
where $X,Y$ represent the input images and corresponding labels respectively. $S$ denotes the causal factors that causally influences both $X$ and $Y$, i.e., the category-related information such as "\textit{shape}" in digit recognition, while $U$ denotes the non-causal factors that only causally influence $X$, which is generally  domain-related information such as \textit{"style"}. 
$V_1, V_2$ are unexplained noise
variables that are jointly independent.
As for $f,h,g$, they can be regarded as unknown structural functions.
Thus, for any distributions $P(X,Y) \in \mathcal{P}$, if the causal factors $S$ is given, there exists a general conditional distribution $P(Y \mid S)$, i.e., an invariant causal mechanism.
Based on the discussion above, if we can access the causal factors, then it is easy to obtain the causal mechanism that generalizes well outside i.i.d. assumption by optimizing $h$ :
\begin{equation}\small
\label{eq:scm}
    h^* = \arg \min_h \mathbb{E}_{P} [\ell(h(g(X)),Y)] = \arg \min_h \mathbb{E}_{P} [\ell(h(S),Y)],
\end{equation}
where $\ell(\cdot,\cdot)$ is the cross entropy loss.

Unfortunately, the causal factors $S$ are not provided to us a priori, we are given raw images $X$ instead which are generally unstructured. It is somehow impractical to directly reconstruct the causal factors as well as mechanisms since they are unobservable and ill-defined. What's more, as discussed by~\cite{B2016Causal, TCRL,LocatelloPRSBT20}, 
which factors can be extracted and their granularity depends on which distribution shifts, supervision
signals and explicit interventions are available. 
Nevertheless, what is clear is that the causal factors still need to obey certain requirements. Previous works~\cite{2017Elements,anti-causal}
declare
that causal factors should be jointly independent, as expressed in Principle~\ref{ICM}.
\begin{principle}[\cite{2017Elements,anti-causal}]\label{ICM}
\textbf{Independent Causal Mechanisms (ICM) Principle}:
The conditional distribution of each variable
given its causes (i.e., its mechanism) does not inform
or influence the other mechanisms.
\end{principle}

Since the $S$ in Eq.~\eqref{assignment} represents the set of all causal factors $\{s_1,s_2,\ldots,s_N\}$, the principle tells us that 1) changing (or performing intervention upon) one mechanism $P(s_i\mid PA_i)$ does not change any of the other mechanisms $P(s_j \mid PA_j), i \neq j $~\cite{anti-causal} ($PA_i$ denotes $s_i$'s parents in causal graph, which can be regarded as causal information that $s_i$ contains since $S$ is already the root node),  2) knowing some other mechanisms $P(s_i \mid PA_i)$ does not give us information about a mechanism $P(s_j \mid PA_j)$~\cite{JanzingS10}. 
Thus, we can factorize the joint distribution of causal factors into conditional as follows, which refers to causal factorization:
\begin{equation}\small
    P(s_1,s_2,\ldots,s_N) = \prod \limits_{i=1}^N P(s_i \mid PA_i),
\end{equation}

Therefore, we highlight that the causal factors $S$ should satisfy three basic properties based on the definition of causal variables in Common Cause Principle (Principle~\ref{CCP}) and the nature of causal mechanism in ICM Principle (Principle~\ref{ICM}):

\begin{itemize}
    \item The causal factors $S$ should be separated from the non-causal factors $U$, i.e., $S \Vbar U$. Thus, performing an intervention upon $U$ does not make changes to $S$.
    \item The factorization $s_1,s_2,\ldots,s_N$ should be jointly independent, none of which entails information of others.
    \item The causal factors $S$ should be causally sufficient to the classification task $X \xrightarrow{} Y$, 
    i.e., contain information that can explain all the statistical dependencies.
\end{itemize}
Thus, instead of directly reconstructing the causal factors, we propose to learn causal representations as an alternative by forcing them to have the same properties as causal factors. We will explain the details in Sec.~\ref{CIRL}

\subsection{Causality Inspired Representation Learning}
\label{CIRL}

In this section, we illustrate our proposed representation learning algorithm inspired by the causality discussed above, which consists of three modules: causal intervention module, causal factorization module and adversarial mask module. The whole framework is shown in Fig.~\ref{fig:framework}.

\subsubsection{Causal Intervention Module}
\label{Sec:CInt}

We first aim to separate the causal factors $S$ from the mixture of non-causal ones $U$ by causal intervention.
Specifically, although the explicit form of causal factor extractor $g(\cdot)$ in Eq.~\eqref{eq:scm} is unknown in general, we have prior knowledge that the causal factors $S$ should remain invariant to the intervention upon $U$, i.e., $P(S\mid do(U))$. 
While in DG literature, we do know that some domain-related information cannot determine the category of inputs, which can be regarded as non-causal factors and captured by some techniques~\cite{FACT,MixStyle,L2A-OT}.
For instance, Fourier transformation has a well-known property: the phase component of Fourier spectrum preserves high-level semantics of the original signal, while the amplitude component contains low-level statistics~\cite{1981fourier,1982fourier}. Thus, we conduct the intervention upon $U$ by disturbing the amplitude information while keeping the phase information unchanged as~\cite{FACT} does.
Formally, given an original input image $\boldsymbol{x}^o$, its Fourier transformation can be formulated as :
\begin{equation}\small
    \mathcal{F}(\boldsymbol{x}^o) = \mathcal{A}(\boldsymbol{x}^o) \times e^{-j \times \mathcal{P}(\boldsymbol{x}^o)},
\end{equation}
where $\mathcal{A}(\boldsymbol{x}^o), \mathcal{P}(\boldsymbol{x}^o)$ denote the amplitude and phase components respectively. The Fourier transformation $\mathcal{F}(\cdot)$ and its inverse $\mathcal{F}^{-1}(\cdot)$ can be calculated with the FFT algorithm~\cite{1981Fast} effectively.
We then perturb the amplitude information via linearly interpolating between the amplitude spectrums of
\lv{
the original image $\boldsymbol{x}^o$ and an image $(\boldsymbol{x}')^o$ sampled randomly from arbitrary source domains:
}
\begin{equation}\small
    \hat{\mathcal{A}}(\boldsymbol{x}^o) =  (1-\lambda)\mathcal{A}(\boldsymbol{x}^o) + \lambda\mathcal{A}((\boldsymbol{x}')^{o}),
\end{equation}
where $\lambda \sim U(0,\eta)$ and $\eta$ controls the strength of perturbation. Then we combine the perturbed amplitude spectrums with the original phase component to generate the augmented image $\boldsymbol{x}^a$ by inverse Fourier transformation :
\begin{equation}\small
\label{fourier_trans}
\mathcal{F}(\boldsymbol{x}^a) = \hat{\mathcal{A}}(\boldsymbol{x}^o) \times e^{-j \times \mathcal{P}(\boldsymbol{x}^o)},\boldsymbol{x}^a = \mathcal{F}^{-1}(\mathcal{F}(\boldsymbol{x}^a)).
\end{equation}

Denote the representation generator implemented by a CNN model as $\hat{g}(\cdot)$ and the representations as $\rb = \hat{g}(\boldsymbol{x}) \in \mathbb{R}^{1\times N}$, where $N$ is the number of dimensions.
To simulate the causal factors that remain invariant to the intervention upon $U$, we optimize $\hat{g}$ to enforce the representations to keep unchanged dimension-wisely to the above intervention:
\begin{equation}\small
\label{CInt}
       \max_{\hat{g}} \frac{1}{N}\sum_{i=1}^{N}COR(\tilde{\boldsymbol{r}}_i^o,\tilde{\boldsymbol{r}}_i^a),
\end{equation}
where $\tilde{\boldsymbol{r}}_i^o$ and $\tilde{\boldsymbol{r}}_i^a$ denote the Z-score normalized $i$-th column of $\Rb^o=[(\rb_1^o)^T,\ldots,(\rb_B^o)^T]^T\in\mathbb{R}^{B\times N}$ and $\Rb^a=[(\rb_1^a)^T,\ldots,(\rb_B^a)^T]^T$, respectively, $B\in\mathbb{Z}_+$ is the batch size, $\rb_i^o = \hat{g}(\boldsymbol{x}_i^o)$ and $\rb_i^a = \hat{g}(\boldsymbol{x}_i^a)$ for $i\in\{1,\ldots,B\}$. We leverage a $COR$ function to measure the correlation of representations before and after the intervention
. Thus, we can realize the first step of simulating causal factors $S$ with representations $R$ by making them independent of $U$.

\subsubsection{Causal Factorization Module}
\label{sec: CFac Module}

As we proposed in Sec.~\ref{subsec:sec_31}
that the factorization of causal factors $s_1,s_2\ldots,s_N$ should be jointly independent in the sense of none of them entails information of others. Therefore, we intend to make any two dimensions of the representations independent of each other:
\begin{equation}\small
\label{Fac}
     \min_{\hat{g}} \frac{1}{N(N-1)}\sum_{i\neq j}COR(\tilde{\boldsymbol{r}}_i^o,\tilde{\boldsymbol{r}}_j^a), i\neq j,
\end{equation}
Note that for saving computation cost, we omit the constraints within $\Rb^o$ or $\Rb^a$.
To unify the optimization goals of Eq.~\eqref{CInt} and Eq.~\eqref{Fac}, we build a correlation matrix $\boldsymbol{C}$:
\begin{equation}\small
\label{CorrelationMatrix}
\boldsymbol{C}_{ij} = \frac{<\tilde{\boldsymbol{r}}_i^o,\tilde{\boldsymbol{r}}_j^a>}{\lVert \tilde{\boldsymbol{r}}_i^o \rVert \lVert \tilde{\boldsymbol{r}}_j^a \rVert}, i,j \in {1,2,\ldots,N},
\end{equation}
where $<\cdot>$ denotes the inner product operation.
Thus, the same dimension of $\Rb^o$ and $\Rb^a$ can be taken as positive pairs which need to maximize the correlation, while the different dimensions can be taken as negative pairs which need to minimize the correlation. Based on that, we design a Factorization loss $\mathcal{L}_{Fac}$ which can be formulated as follows:
\begin{equation}\small
\label{L_fac}
\mathcal{L}_{Fac} = \frac{1}{2}\lVert \boldsymbol{C} - \boldsymbol{I} \rVert _F^2.
\end{equation}

\textbf{Remark 1.}
The objective in Eq.~\eqref{L_fac} can make the diagonal elements of correlation matrix $\boldsymbol{C}$ approximate to 1, which means that the representations before and after the intervention to non-causal factors are invariant. It indicates that we can effectively separate the causal factors from the mixture of the non-causal ones. Moreover, it also makes the non-diagonal elements of $\boldsymbol{C}$ close to 0, i.e., enforcing the dimensions of representations to be jointly independent. 
Thus, by minimizing $\mathcal{L}_{Fac}$, we can make the noisy and dependent representations into clean and independent ones, satisfying the first two properties of ideal causal factors.

\subsubsection{Adversarial Mask Module}

To succeed on the classification task $X \xrightarrow{} Y$, the representations should be causally sufficient that entails all the support information. 
The most straightforward way is to utilize the supervision labels $y$ in the multiple source domains:
\begin{equation}\small
\label{CE}
\mathcal{L}_{cls} = \ell(\hat{h}(\hat{g}(\boldsymbol{x}^o)),y) + \ell(\hat{h}(\hat{g}(\boldsymbol{x}^a)),y)
\end{equation}
where $\hat{h}$ is the classifier. However, this straightforward way
cannot guarantee that every dimension of our learned representations are important, i.e., contain sufficient underlying causal information for classification. Specifically, there may be inferior dimensions that carry relatively less causal information and then make a small contribution for classification. Therefore, we propose to detect these dimensions and enforce them to contribute more. Since the dimensions are also required to be jointly independent with the aid of our factorization module, the detected inferior dimensions are rendered to contain more and novel causal information that is not included by other dimensions, which makes the whole representations to be more causality sufficient.

Thus, to detect the inferior dimensions, we design an adversarial mask module.
We build a neural-network-based masker, denoted by $\hat{w}$
to learn the contribution of each dimension, and the dimensions correspond to the largest $\kappa\in(0,1)$ ratio are regarded as superior dimensions while the rest are regarded as the inferior ones:
\begin{equation}\small
\label{masker}
\boldsymbol{m} = \mbox{Gumbel-Softmax}(\hat{w}(\rb), \kappa N) \in \mathbb{R}^N,\\
\end{equation}
where we employ the commonly-used derivable Gumbel-Softmax trick~\cite{gumbel} to sample a mask with $\kappa N$ values approaching $1$. The details of the trick are deferred to supplementary materials.
By multiplying the learned representations by the obtained masks $\boldsymbol{m}$ and $1-\boldsymbol{m}$, we can acquire the superior and inferior dimensions of the representations, respectively.
Then, we feed them into two different classifiers $\hat{h}_1, \hat{h}_2$. Eq.~\eqref{CE} can be rewritten as follows:
\begin{equation}\small
\label{loss_ad}
\begin{split}
  \mathcal{L}_{cls}^{sup} =& \ell(\hat{h}_1(\rb^o\odot \boldsymbol{m}^o),y) + \ell(\hat{h}_1(\rb^a\odot \boldsymbol{m}^a),y),\\ 
  \mathcal{L}_{cls}^{inf} =& \ell(\hat{h}_2(\rb^o\odot (1-\boldsymbol{m}^o)),y)+\ell(\hat{h}_2(\rb^a\odot (1-\boldsymbol{m}^a)),y),
\end{split}
\end{equation}
We optimize the masker by minimizing $\mathcal{L}_{cls}^{sup}$ and maximizing $\mathcal{L}_{cls}^{inf}$, while optimize the generator $\hat{g}$ and classifiers $\hat{h}_1, \hat{h}_2$ by minimizing the two supervision loss.

\textbf{Remark 2.} The proposed adversarial mask module can precisely detect the inferior dimensions because 1) for an optimized $\hat{h}_2$ to minimize $\mathcal{L}_{cls}^{inf}$ based on existing masked dimensions, learning $\boldsymbol{m}$ to select dimensions for maximizing $\mathcal{L}_{cls}^{inf}$ can find inferior dimensions with less contribution, and 2) superior and inferior dimension-sets complement each other such that if one dimension is not treated as superior then it will be treated as inferior, thus the selection of the superior ones will help the selection of the inferior ones. Furthermore, compared with optimizing Eq.~\eqref{CE} only, the adversarial mask module combined with our causal factorization module can help to generate more causally sufficient representations, because by optimizing $\hat{g}$ to minimize both $\mathcal{L}_{cls}^{inf}$ and $\mathcal{L}_{Fac}$, the inferior dimensions are forced to carry more causal information and be independent with existing superior dimensions. Finally, the learned representations will approach causal sufficient by iteratively ``replacing'' the inferior representations to be novel superior ones.


To be clear, the overall optimization objective of our proposed CIRL is summarized as follows:
\begin{equation}\small
\label{optimization}
\begin{split}
    &\min_{\hat{g},\hat{h}_1, \hat{h}_2}\mathcal{L}_{cls}^{sup} +\mathcal{L}_{cls}^{inf} + \tau \mathcal{L}_{Fac}, \quad \min_{\hat{w}} \mathcal{L}_{cls}^{sup} -\mathcal{L}_{cls}^{inf},
\end{split}
\end{equation}
where $\tau$ is the trade-off parameter. Note that the whole representation $\rb$ and classifier $\hat{h}_1$ are utilized during inference.


\lv{\textbf{Remark 3.} Note that the effect of the number of feature dimension is negligible. Through the cooperative optimization of the three modules, the total amount of causal information contained in the whole representations will increase until the learned representations can explain all the statistical dependence between inputs and labels, regardless of the feature dimension. Experimental analyses are provided in supplementary materials, verifying our demonstration.
}
\section{Experiment}
\label{sec:exp}

\subsection{Datasets}

\textbf{Digits-DG} \cite{Digits-DG} encompasses four digit domains including \textit{MNIST}\cite{MNIST}, \textit{MNIST-M}\cite{SYN_MNIST-M}, \textit{SVHN} \cite{SVHN} and \textit{SYN} \cite{SYN_MNIST-M}, which present dramatic differences in the font style, background and stroke color. Following \cite{Digits-DG}, we randomly select $600$ images per class for each domain, and then split $80\%$ of the data for training and $20\%$ of the data for validation.

\textbf{PACS} \cite{DBA} is specially proposed for DG, which contains $9,991$ images from four domains (\textit{Art-Painting}, \textit{Cartoon}, \textit{Photo} and \textit{Sketch}) with large style discrepancy. In each domain, there are $7$ categories: dog, elephant, giraffe, guitar, house, horse, and person. For a fair comparison, the original training-validation split provided by \cite{DBA} is used.

\textbf{Office-Home} \cite{Office-Home} is an object recognition dataset in the office and home environments, which collects $15,500$ images of $65$ categories. The $65$ categories are shared by four domains (\textit{Art}, \textit{Clipart}, \textit{Product} and \textit{Real-World}), which differs in the viewpoint and image style. Following \cite{FACT}, each domain is split into $90\%$ for training and $10\%$ for validation.

\subsection{Implementation Details}
\lv{
Following the commonly used leave-one-domain-out protocol \cite{DBA}, we specify one domain as the unseen target domain for evaluation and train with the remaining domains.
For Digits-DG, all images are resized to $32\times32$, we train the network from scratch using the mini-batch SGD optimizer with batch size $128$, momentum $0.9$, and weight decay $5e$-$4$ for $50$ epochs. And the learning rate is decayed by $0.1$ every $20$ epochs.
As for PACS and Office-Home, all images are resized to $224\times 224$. 
The network is trained from scratch using the mini-batch SGD with batch size $16$, momentum $0.9$ and weight decay $5e$-$4$ for $50$ epochs, and the learning rate is decayed by $0.1$ at $80\%$ of the total epochs. 
For the hyper-parameters $\kappa$ and $\tau$, their values are selected according to the results on the source validation set, since the target domain is unseen during the training. Specifically, we set $\kappa = 60\%$ for Digits-DG and PACS while $\kappa = 80\%$ for Office-Home. $\tau$ is set as $2$ for Digits-DG and $5$ for the others.
All results are reported based on the average accuracy over three repetitive runs.
More details are given in the supplementary details.
}


\subsection{Experimental Results}

\textbf{Results on Digits-DG} are presented in Table \ref{digitsdg}, where CIRL beats all the compared baselines in terms of the average accuracy. Note that CIRL surpasses the domain-invariant representation based methods CCSA \cite{CCSA} and MMD-AAE \cite{MMD-AAE} by a large margin of $8.0\%$ and $7.9\%$, respectively, which indicates the importance of excavating the intrinsic causal mechanisms between data and labels, instead of the superficial statistical dependence. Besides, we also compare CIRL with FACT \cite{FACT}, since our causal intervention module adopts the same augmentation technique. It is worth mentioning that FACT is a quite state-of-the-art method in DG community and $1.0\%$ performance improvement is challenging. While CIRL achieves $1.0\%$ improvement over FACT, which further validates the effectiveness of our method.

\begin{table}[!t]
\setlength{\abovecaptionskip}{0.cm}
\setlength{\belowcaptionskip}{0.cm}
\caption{Leave-one-domain-out results on Digits-DG. 
The best and second-best results are bold and underlined, respectively. 
}
  \centering
    \setlength{\tabcolsep}{0.8mm}{
    \begin{tabular}{l|cccc|c}
    \toprule
    Methods & MNIST & MNIST-M & SVHN & SYN & Avg. \\
    \midrule
    DeepAll~\cite{Digits-DG} & 95.8 & 58.8 & 61.7 & 78.6 & 73.7 \\
    Jigen~\cite{Jigen} & 96.5& 61.4 & 63.7 & 74.0 & 73.9 \\
    CCSA~\cite{CCSA} & 95.2 & 58.2 & 65.5 & 79.1 & 74.5 \\
    MMD-AAE~\cite{MMD-AAE} & 96.5 & 58.4 & 65.0 & 78.4 & 74.6 \\
    CrossGrad~\cite{CrossGrad} & \underline{96.7} & 61.1 & 65.3 & 80.2 & 75.8 \\
    DDAIG~\cite{Digits-DG} & 96.6 & 64.1 & 68.6 & 81.0 & 77.6 \\
    L2A-OT~\cite{L2A-OT} & \underline{96.7} & 63.9 &68.6& 83.2 & 78.1 \\
    FACT \cite{FACT} & \textbf{97.9} & \underline{65.6} & \underline{72.4} & \textbf{90.3} & \underline{81.5} \\
    \midrule
    CIRL (\textit{ours}) & 96.08 &	\textbf{69.87}&	\textbf{76.17}&	\underline{87.68}&	\textbf{82.5} \\
    \bottomrule
    \end{tabular}}
  \label{digitsdg}
  \vspace{-3mm}
\end{table}

\begin{table}[!t]\small
  \setlength{\abovecaptionskip}{0.cm}
  \setlength{\belowcaptionskip}{0.cm}
  \centering
  \caption{Leave-one-domain-out results on PACS with ResNet-18. 
  }
    \resizebox{\columnwidth}{!}{\begin{tabular}{l|cccc|c}
    \toprule
    Methods & Art & Cartoon & Photo & Sketch & Avg. \\
    \midrule
    DeepAll\cite{Digits-DG} & 77.63 & 76.77 & 95.85 & 69.50 & 79.94 \\
    MetaReg~\cite{MetaReg} & 83.70& 77.20 & 95.50 & 70.30 & 81.70 \\
    JiGen~\cite{Jigen} & 79.42 & 75.25 & 96.03 & 71.35 & 80.51 \\
    DDAIG~\cite{Digits-DG} & 84.20 & 78.10 & 95.30 & 74.70 & 83.10 \\
    CSD~\cite{CSD} & 78.90 & 75.80 & 94.10 & 76.70 & 81.40 \\
    MASF~\cite{MASF} & 80.29 & 77.17 & 94.99 & 71.69 & 81.04 \\
    L2A-OT~\cite{L2A-OT} & 83.30 & 78.20 & 96.20 & 73.60 & 82.80 \\
    EISNet~\cite{EISNet} & 81.89 & 76.44 & 95.93 & 74.33 & 82.15 \\
    MatchDG \cite{MatchDG} & 81.32 & \textbf{80.70} & \underline{96.53} & 79.72 & 84.56 \\
    RSC~\cite{RSC} & 83.43 & 80.31 & 95.99 & 80.85 & 85.15 \\
    FACT \cite{FACT}& \underline{85.90}&	79.35	&\textbf{96.61}	&\underline{80.88}&	\underline{85.69} \\
    \midrule
    CIRL (\textit{ours})  & \textbf{86.08} & \underline{80.59} & 95.93 & \textbf{82.67} & \textbf{86.32} \\
    \bottomrule
    \end{tabular}}
  \label{pacs-res18}
 \vspace{-3mm}
\end{table}

\begin{table}[!t]\small
  \setlength{\abovecaptionskip}{0.cm}
  \setlength{\belowcaptionskip}{0.cm}
  \centering
  \caption{Leave-one-domain-out results on PACS with ResNet-50. 
  }
    \resizebox{\columnwidth}{!}{
    \begin{tabular}{l|cccc|c}
    \toprule
    Methods & Art & Cartoon & Photo & Sketch & Avg. \\
    \midrule
    DeepAll\cite{Digits-DG} & 84.94 & 76.98 & 97.64 & 76.75 & 84.08 \\
    MetaReg~\cite{MetaReg} & 87.20 & 79.20 & 97.60 & 70.30 & 83.60 \\
    MASF~\cite{MASF}  & 82.89 & 80.49 & 95.01 & 72.29 & 82.67 \\
    EISNet~\cite{EISNet} & 86.64 & 81.53 & 97.11 & 78.07 & 85.84 \\
    MatchDG \cite{MatchDG} & 85.61 & 82.12 & \textbf{97.94} & 78.76 & 86.11 \\
    FACT \cite{FACT} & \textbf{90.89} & \underline{83.65} & 97.78 & \underline{86.17} & \underline{89.62} \\
    \midrule
    CIRL (\textit{ours}) & \underline{90.67} & \textbf{84.30} & 97.84 & \textbf{87.68} & \textbf{90.12} \\
    \bottomrule
    \end{tabular}}
  \label{pacs-50}
    \vspace{-3mm}
\end{table}

\begin{table}[!t]\small
  \setlength{\abovecaptionskip}{0.cm}
  \setlength{\belowcaptionskip}{0.cm}
  \centering
  \caption{Leave-one-domain-out results on Office-Home
}
    \resizebox{\columnwidth}{!}{
    \begin{tabular}{l|cccc|c}
    \toprule
    Methods & Art & Clipart & Product & Real & Avg. \\
    \midrule
    DeepAll \cite{Digits-DG} & 57.88 & 52.72 & 73.50 & 74.80 & 64.72 \\
    CCSA~\cite{CCSA}  & 59.90 & 49.90 & 74.10 & 75.70 & 64.90 \\
    MMD-AAE~\cite{MMD-AAE} & 56.50 & 47.30 & 72.10 & 74.80 & 62.70 \\
    CrossGrad~\cite{CrossGrad} & 58.40 & 49.40 & 73.90 & 75.80 & 64.40 \\
    DDAIG~\cite{Digits-DG} & 59.20 & 52.30 & 74.60 & 76.00 & 65.50 \\
    L2A-OT~\cite{L2A-OT} & \underline{60.60} & 50.10 & \underline{74.80} & \textbf{77.00} & 65.60 \\
    Jigen~\cite{Jigen} & 53.04 & 47.51 & 71.47 & 72.79 & 61.20 \\
    RSC~\cite{RSC}   & 58.42 & 47.90 & 71.63 & 74.54 & 63.12 \\
    FACT \cite{FACT}& 60.34 & \underline{54.85} & 74.48 & 76.55 & \underline{66.56} \\
    \midrule
     CIRL (\textit{ours}) & \textbf{61.48} & \textbf{55.28} & \textbf{75.06} & \underline{76.64} & \textbf{67.12} \\
    \bottomrule
    \end{tabular}}
  \label{officehome}
      \vspace{-6mm}
\end{table}

\textbf{Results on PACS} based on ReNet-18 and ResNet-50 are reported in Table \ref{pacs-res18} and \ref{pacs-50}, respectively. It can be observed that CIRL obtains the highest average accuracy among all the compared methods on both backbones. Specifically, compared with MatchDG \cite{MatchDG}, which also introduces causality into DG problem, CIRL outperforms MatchDG by a large margin of $1.76\%$ on ResNet-18 and $4.01\%$ on ResNet-50. Because CIRL explicitly learns causal representations based on a more theoretical formulation, instead of in an implicit manner. 
\lv{There also exist cases where CIRL performs relatively poorly, this may be due to the tasks being fairly saturated in performance such as the \textit{photo} task, or due to bad image qualities like noise samples which contain damaged causal information. However, we still achieve the second-best on these tasks and our overall performance outperforms others.}
In general, the encouraging results demonstrate the superiority of our causal reconstruction technique in CIRL.

\textbf{Results on Office-Home} based on ResNet-18 are summarized in Table \ref{officehome}. The larger number of categories and samples makes Office-Home a more challenging benchmark than PACS for domain generalization. On this challenging benchmark, CIRL still achieves the best average performance of $67.12\%$, surpassing FACT \cite{FACT} by a margin of $0.56\%$. The improvements further justify the efficacy of CIRL.

\subsection{Analytical Experiments}
\textbf{Ablation Study.} We discuss the influences of the Causal Intervention (\textit{CInt.}) module, Causal Factorization (\textit{CFac.}) module and Adversarial Mask (\textit{AdvM.}) module in CIRL. Table \ref{ablation} presents the results of different variants of CIRL on PACS dataset with ResNet-18 as the backbone. Comparing variants 1, 2 with variant 3, we can observe that the performance of combining both CInt. and CFac. modules is much better, which indicates that only separating the representation from non-causal factors or making the dimensions independent is not sufficient for well modeling the causal factors. Besides, the improved performance of variant 4 over variant 2 implies that the AdvM. module contributes to integrating more informative information into representations for classification. Finally, CIRL performs best, showing that the three modules complement and promote mutually, none of which is indispensable for the superior generalization ability.

 \begin{figure*}[t]
  \setlength{\abovecaptionskip}{0.cm}
  \setlength{\belowcaptionskip}{0.cm}
  \centering
  \includegraphics[width=1.0\linewidth]{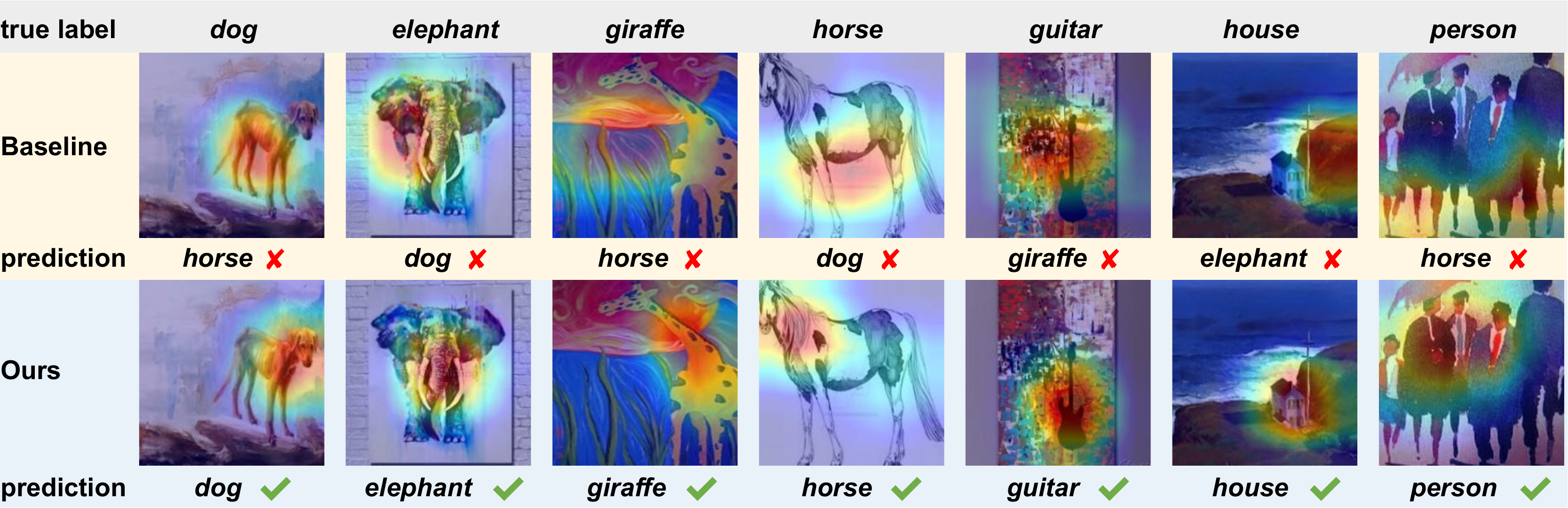}
   \caption{Visualization of attention maps of the last convolutional layer on PACS dataset, with Art-Painting as the unseen target domain.}
   \label{fig:scml}
   \vspace{-4mm}
\end{figure*}

\begin{figure}
  \centering
  \begin{subfigure}{0.48\linewidth}
    \includegraphics[width=1.0\linewidth]{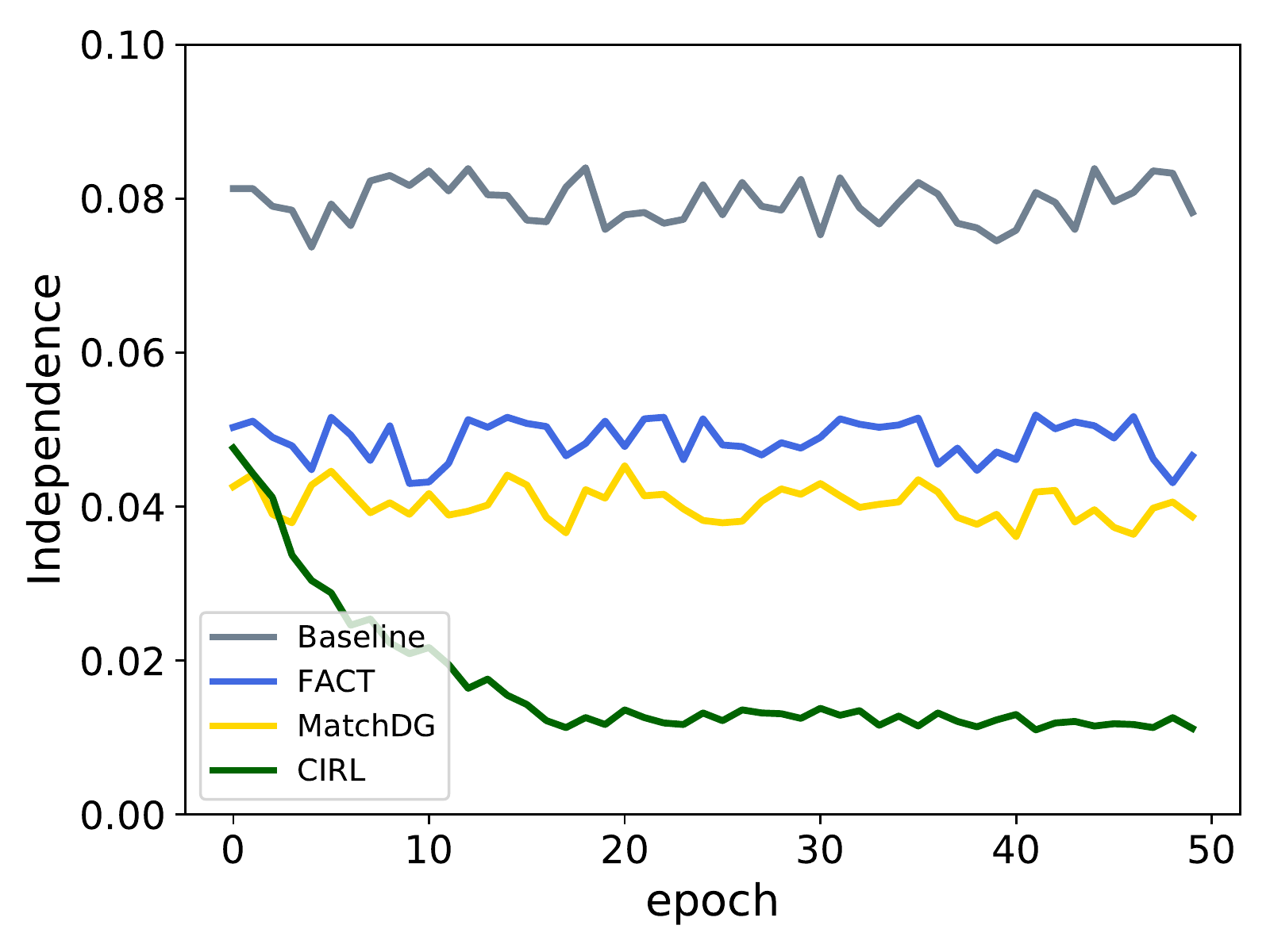}
    \caption{ResNet-18.}
    \label{fig:ind-18}
  \end{subfigure}
  \hfill
  \begin{subfigure}{0.48\linewidth}
  \includegraphics[width=1.0\linewidth]{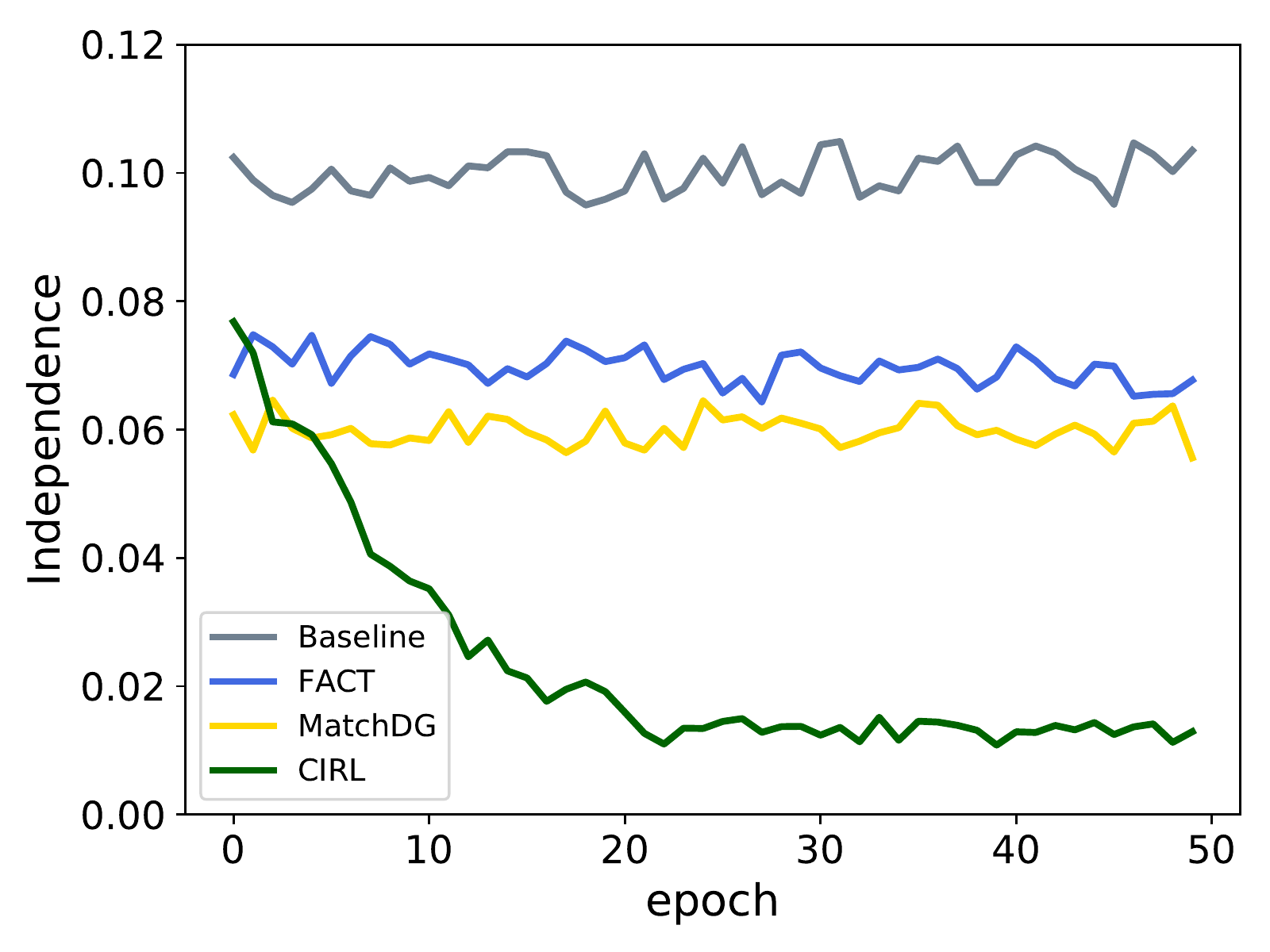}
    \caption{ResNet-50.}
    \label{fig:ind-50}
  \end{subfigure}
  \hfill
  \begin{subfigure}{0.48\linewidth}
    \includegraphics[width=1.0\linewidth]{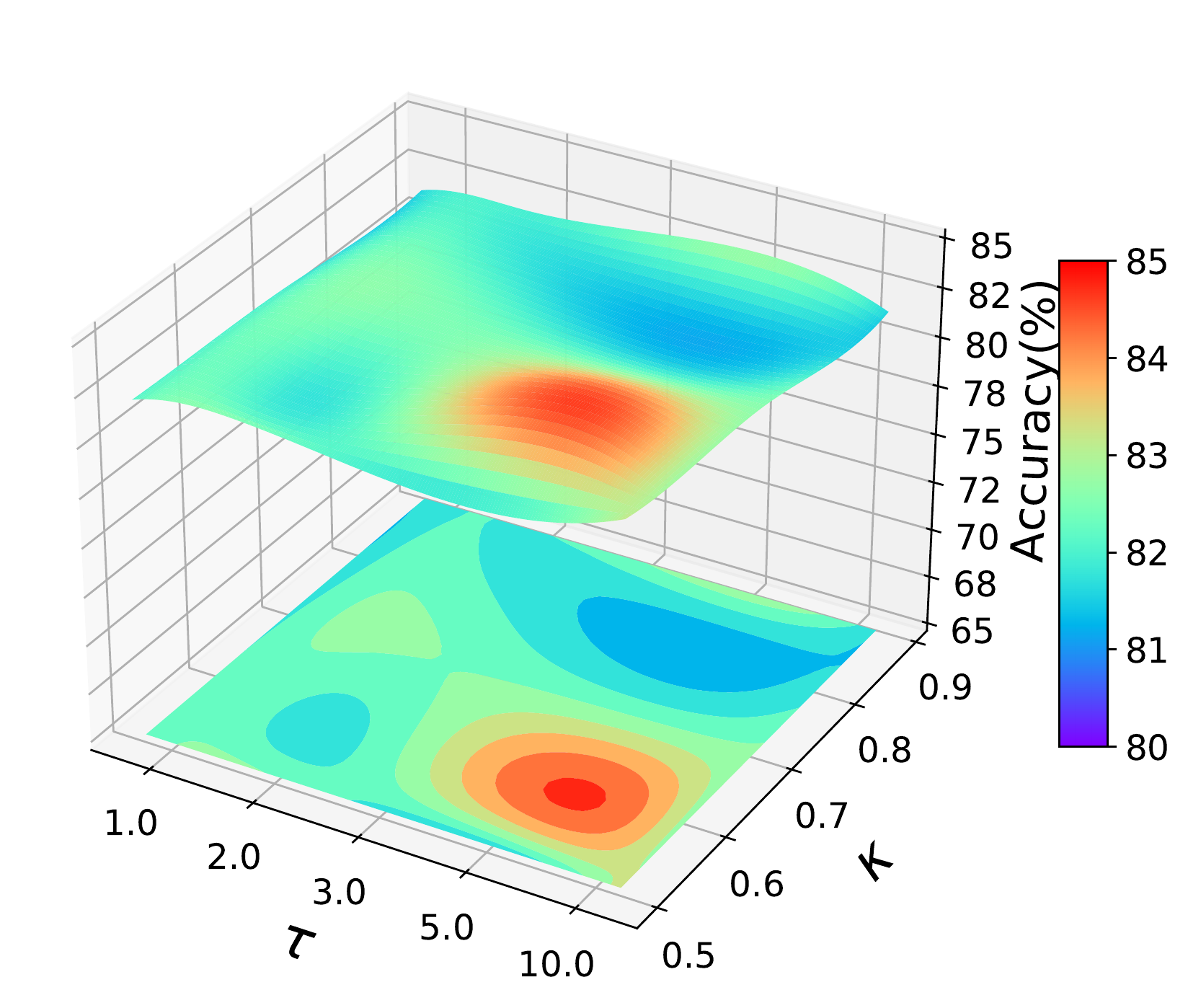}
    \caption{ResNet-18.}
    \label{fig:sens-18}
  \end{subfigure}
  \hfill
  \begin{subfigure}{0.48\linewidth}
  \includegraphics[width=1.0\linewidth]{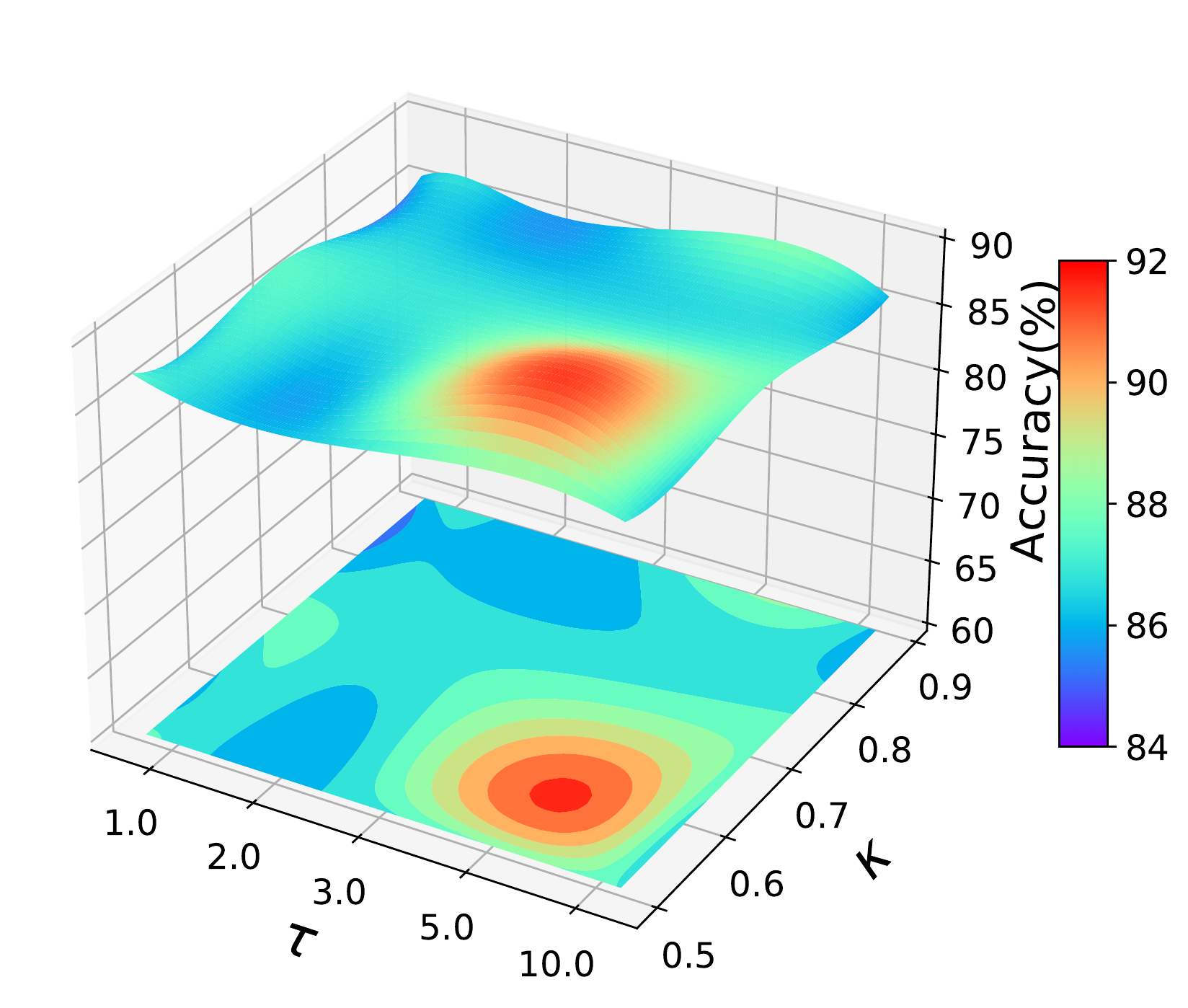}
    \caption{ResNet-50.}
    \label{fig:sens-50}
  \end{subfigure}
  \vspace{-2mm}
  \caption{(a) and (b) are the Independence Degree, (c) and (d) are the sensitivity of CIRL to hyper-parameters $\tau$ and $\kappa$. All results are conducted on PACS dataset with Sketch as unseen target domain.}
  \label{fig:CwithP}
  \vspace{-2mm}
\end{figure}

\textbf{Visual Explanation.} To visually verify the claim that the representations learned by CIRL can simulate the causal factors, we provide the attention maps of the last convolutional layer for baseline (i.e., DeepAll) and CIRL methods, utilizing the visualization technique in \cite{Grad-CAM}. The results are shown in Fig. \ref{fig:scml}. We can see that the representations learned by CIRL are more category-related, compared with the baseline method. Taking the giraffe for example, its long neck can be regarded as one of the causal factors to classify, which is precisely captured by CIRL. While the baseline focuses on non-causal factors (e.g., the texture resembling the mane), which leads to the misclassification. More visualization results are given in the supplementary materials.

\begin{table}
\small
  \setlength{\abovecaptionskip}{0.cm}
  \setlength{\belowcaptionskip}{0.cm}
\caption{Ablation study of CIRL on PACS dataset with ResNet-18.}
  \centering
    \setlength{\tabcolsep}{0.1mm}{
    \begin{tabular}{c|ccc|cccc|c}
    \toprule
    Method & CInt. & CFac. & AdvM. & Art & Cartoon & Photo & Sketch & Avg.\\
    \midrule
    Variant 1 & $\checkmark$ & - & - & 85.99 &	78.67	&95.81&	79.49&	84.99 \\
    Variant 2 & - & $\checkmark$ & - & 80.52 &78.75	&94.37&	78.57&	83.05 \\
    Variant 3 & $\checkmark$ & $\checkmark$ & - & 84.23&79.44&	95.51&	82.54&	85.43 \\
    Variant 4 & - & $\checkmark$ & $\checkmark$ & 82.18&79.69&	95.99&	80.68&	84.64 \\
    CIRL & $\checkmark$ & $\checkmark$ & $\checkmark$ & 86.08 & 80.59 & 95.93 & 82.67 & \textbf{86.32} \\
    \bottomrule
    \end{tabular}}
  \label{ablation}
  \vspace{-4mm}
\end{table}

\textbf{Independence of Causal Representation.} Fig. \ref{fig:ind-18} and \ref{fig:ind-50} show the independence degree between dimensions of representations on task Sketch. Specifically, we use $\|\boldsymbol{C}\|_{F}^2 - \|diag(\boldsymbol{C})\|_{2}^2$ as metric and smaller value denotes better independence degree, where $\boldsymbol{C}$ is the correlation matrix in Section \ref{sec: CFac Module}. It is clear that on both ResNet-18 and ResNet-50, the baseline methods have poor independence while the independence of CIRL increases with the training epochs and finally achieves a high degree with stabilization. The results demonstrate the efficacy of our designed causal factorization module, which allows our learned representations can be served as the mimic of causal factors.


\begin{table}[!t]\small
  \setlength{\abovecaptionskip}{0.cm}
  \setlength{\belowcaptionskip}{0.cm}
\caption{Comparisons of the representation importance for different methods on PACS dataset. [Std. denotes the standard deviation.]}
  \centering
  \setlength{\tabcolsep}{0.8mm}{\begin{tabular}{c|ccccc}
    \toprule
   
              Methods & Mean & Std.  \\
    \midrule
     DeepAll \cite{Digits-DG} &  3.10 & 0.25 \\
     FACT \cite{FACT} &  3.12 & 0.28  \\
     MatchDG \cite{MatchDG} &  3.17 & 0.19  \\
     CIRL (\textit{ours}) & \textbf{3.22} & \textbf{0.09}\\
    \bottomrule
    \end{tabular}}
  \label{importance}
  \vspace{-4mm}
\end{table}
\textbf{Representation Importance.} Ideally, we hope each dimension of the representations can be important that involves as much and as conducive causal information as possible, making the whole representations causally sufficient for classification. Inspired by \cite{HassanpourG20}, we exploit the weights for the first layer of the classifier to estimate such importance of each representation dimension. Note that we conduct normalization on the weights for each dimension, i.e., $(x - min) / (max - min)$ to deal with the scale issue. We show the mean and standard deviation of the importance for all dimensions in Table \ref{importance} since the number of dimensions is large.
It can be observed that MatchDG \cite{MatchDG} and CIRL present higher representation importance since they embed the causal information that truly affects the classification into representations. Moreover, the highest mean value and the lowest standard deviation that CIRL achieves indicate that each dimension of our learned representations is important, which further shows the superiority of CIRL.

\textbf{Parameter Sensitivity.} Fig. \ref{fig:sens-18} and \ref{fig:sens-50} show the sensitivity of CIRL to hyper-parameters $\tau$ and $\kappa$. Specifically, the value of $\tau$ varies from $\{1.0, 2.0, 3.0, 5.0, 10.0\}$, while $\kappa$ changes from $\{0.5, 0.6, 0.7, 0.8, 0.9\}$. It can be observed that CIRL achieves competitive performances robustly under a wide range of hyper-parameter values, i.e., $5.0 \leq \tau \leq 10.0$ and $0.5 \leq \kappa \leq 0.6$, with either ResNet-18 or ResNet-50 as backbone, which further verify the stability of our method.

\section{Conclusion}
\label{sec:conclusion}


\lv{
This paper points out the insufficiency of statistical knowledge and presents a causal view of DG. The main idea is to reconstruct causal factors and excavate intrinsic causal mechanisms. We then propose a framework CIRL to learn causal representations that can mimic the causal factors based on the ideal properties we highlight.
Comprehensive experiments demonstrate the effectiveness and superiority of CIRL.
Considering the mainstream of related work is generally based on statistical dependence between variables, we hope our work can shed some lights into the community.
}



\lv{\paragraph{Acknowledgements.} This work was supported by the National Natural Science Foundation of China under Grant No. U21A20519 and No. 61902028.}

\bibliographystyle{ieee_fullname}
\bibliography{egbib}

\begin{thebibliography}{10}\itemsep=-1pt

\bibitem{Arjovsky2019}
Mart{\'{\i}}n Arjovsky, L{\'{e}}on Bottou, Ishaan Gulrajani, and David
  Lopez{-}Paz.
\newblock Invariant risk minimization.
\newblock {\em CoRR}, abs/1907.02893, 2019.

\bibitem{MetaReg}
Yogesh Balaji, Swami Sankaranarayanan, and Rama Chellappa.
\newblock Metareg: Towards domain generalization using meta-regularization.
\newblock In {\em NeurIPS}, pages 1006--1016, 2018.

\bibitem{MTL}
Gilles Blanchard, Aniket~Anand Deshmukh, {\"{U}}r{\"{u}}n Dogan, Gyemin Lee,
  and Clayton Scott.
\newblock Domain generalization by marginal transfer learning.
\newblock {\em J. Mach. Learn. Res.}, 22:2:1--2:55, 2021.

\bibitem{CRLS}
L{\'{e}}on Bottou, Jonas Peters, Joaquin~Qui{\~{n}}onero Candela, Denis~Xavier
  Charles, Max Chickering, Elon Portugaly, Dipankar Ray, Patrice~Y. Simard, and
  Ed Snelson.
\newblock Counterfactual reasoning and learning systems: the example of
  computational advertising.
\newblock {\em J. Mach. Learn. Res.}, 14(1):3207--3260, 2013.

\bibitem{Jigen}
Fabio~Maria Carlucci, Antonio D'Innocente, Silvia Bucci, Barbara Caputo, and
  Tatiana Tommasi.
\newblock Domain generalization by solving jigsaw puzzles.
\newblock In {\em CVPR}, pages 2229--2238, 2019.

\bibitem{ChangZYJ20}
Shiyu Chang, Yang Zhang, Mo Yu, and Tommi~S. Jaakkola.
\newblock Invariant rationalization.
\newblock In {\em ICML}, pages 1448--1458, 2020.

\bibitem{chen2018learning}
Jianbo Chen, Le Song, Martin Wainwright, and Michael Jordan.
\newblock Learning to explain: An information-theoretic perspective on model
  interpretation.
\newblock In {\em International Conference on Machine Learning}, pages
  883--892, 2018.

\bibitem{ImageNet}
Jia Deng, Wei Dong, Richard Socher, Li{-}Jia Li, Kai Li, and Li Fei{-}Fei.
\newblock Imagenet: {A} large-scale hierarchical image database.
\newblock In {\em CVPR}, pages 248--255, 2009.

\bibitem{MASF}
Qi Dou, Daniel~Coelho de Castro, Konstantinos Kamnitsas, and Ben Glocker.
\newblock Domain generalization via model-agnostic learning of semantic
  features.
\newblock In {\em NeurIPS}, pages 6447--6458, 2019.

\bibitem{SYN_MNIST-M}
Yaroslav Ganin and Victor~S. Lempitsky.
\newblock Unsupervised domain adaptation by backpropagation.
\newblock In {\em ICML}, pages 1180--1189, 2015.

\bibitem{SCA}
Muhammad Ghifary, David Balduzzi, W.~Bastiaan Kleijn, and Mengjie Zhang.
\newblock Scatter component analysis: {A} unified framework for domain
  adaptation and domain generalization.
\newblock {\em IEEE TPAMI}, 39(7):1414--1430, 2017.

\bibitem{MTA}
Muhammad Ghifary, W.~Bastiaan Kleijn, Mengjie Zhang, and David Balduzzi.
\newblock Domain generalization for object recognition with multi-task
  autoencoders.
\newblock In {\em ICCV}, pages 2551--2559, 2015.

\bibitem{2016Causal}
Madelyn Glymour, Judea Pearl, and Nicholas~P Jewell.
\newblock {\em Causal inference in statistics: A primer}.
\newblock John Wiley \& Sons, 2016.

\bibitem{HassanpourG20}
Negar Hassanpour and Russell Greiner.
\newblock Learning disentangled representations for counterfactual regression.
\newblock In {\em ICLR}, 2020.

\bibitem{resnet}
Kaiming He, Xiangyu Zhang, Shaoqing Ren, and Jian Sun.
\newblock Deep residual learning for image recognition.
\newblock In {\em CVPR}, pages 770--778, 2016.

\bibitem{Heinze-DemlM21}
Christina Heinze{-}Deml and Nicolai Meinshausen.
\newblock Conditional variance penalties and domain shift robustness.
\newblock {\em Mach. Learn.}, (2):303--348, 2021.

\bibitem{imagenet-c}
Dan Hendrycks and Thomas~G. Dietterich.
\newblock Benchmarking neural network robustness to common corruptions and
  perturbations.
\newblock In {\em ICLR}, 2019.

\bibitem{RSC}
Zeyi Huang, Haohan Wang, Eric~P. Xing, and Dong Huang.
\newblock Self-challenging improves cross-domain generalization.
\newblock In {\em ECCV}, pages 124--140, 2020.

\bibitem{2015Causal}
G.~W. Imbens and D.~B. Rubin.
\newblock Causal inference for statistics, social, and biomedical sciences:
  Contents.
\newblock 2015.

\bibitem{gumbel}
Eric Jang, Shixiang Gu, and Ben Poole.
\newblock Categorical reparameterization with gumbel-softmax.
\newblock In {\em ICLR}, 2017.

\bibitem{JanzingS10}
Dominik Janzing and Bernhard Sch{\"{o}}lkopf.
\newblock Causal inference using the algorithmic markov condition.
\newblock {\em {IEEE} Trans. Inf. Theory}, 56(10):5168--5194, 2010.

\bibitem{MNIST}
Yann Lecun, Leon Bottou, Y. Bengio, and Patrick Haffner.
\newblock Gradient-based learning applied to document recognition.
\newblock {\em Proc. IEEE}, 86:2278 -- 2324, 1998.

\bibitem{MMD}
Chen{-}Yu Lee, Tanmay Batra, Mohammad~Haris Baig, and Daniel Ulbricht.
\newblock Sliced wasserstein discrepancy for unsupervised domain adaptation.
\newblock In {\em CVPR}, pages 10285--10295, 2019.

\bibitem{1974Causation}
David Lewis.
\newblock Causation.
\newblock {\em The journal of philosophy, 70(17):}, pages 556--567, 1974.

\bibitem{DBA}
Da Li, Yongxin Yang, Yi{-}Zhe Song, and Timothy~M. Hospedales.
\newblock Deeper, broader and artier domain generalization.
\newblock In {\em ICCV}, pages 5543--5551, 2017.

\bibitem{Ave}
Da Li, Yongxin Yang, Yi{-}Zhe Song, and Timothy~M. Hospedales.
\newblock Learning to generalize: Meta-learning for domain generalization.
\newblock In {\em AAAI}, pages 3490--3497, 2018.

\bibitem{MLDG}
Da Li, Yongxin Yang, Yi{-}Zhe Song, and Timothy~M. Hospedales.
\newblock Learning to generalize: Meta-learning for domain generalization.
\newblock In {\em AAAI}, pages 3490--3497, 2018.

\bibitem{MMD-AAE}
Haoliang Li, Sinno~Jialin Pan, Shiqi Wang, and Alex~C. Kot.
\newblock Domain generalization with adversarial feature learning.
\newblock In {\em CVPR}, pages 5400--5409, 2018.

\bibitem{BCDM}
Shuang Li, Fangrui Lv, Binhui Xie, Chi~Harold Liu, Jian Liang, and Chen Qin.
\newblock Bi-classifier determinacy maximization for unsupervised domain
  adaptation.
\newblock In {\em AAAI}, pages 8455--8464, 2021.

\bibitem{GDCAN}
Shuang Li, Binhui Xie, Qiuxia Lin, Chi~Harold Liu, Gao Huang, and Guoren Wang.
\newblock Generalized domain conditioned adaptation network.
\newblock {\em CoRR}, abs/2103.12339, 2021.

\bibitem{CIAN}
Ya Li, Xinmei Tian, Mingming Gong, Yajing Liu, Tongliang Liu, Kun Zhang, and
  Dacheng Tao.
\newblock Deep domain generalization via conditional invariant adversarial
  networks.
\newblock In {\em ECCV}, pages 647--663, 2018.

\bibitem{FCN}
Yiying Li, Yongxin Yang, Wei Zhou, and Timothy~M. Hospedales.
\newblock Feature-critic networks for heterogeneous domain generalization.
\newblock In {\em ICML}, pages 3915--3924, 2019.

\bibitem{LiuH00S21}
Jiashuo Liu, Zheyuan Hu, Peng Cui, Bo Li, and Zheyan Shen.
\newblock Heterogeneous risk minimization.
\newblock In {\em ICML}, pages 6804--6814, 2021.

\bibitem{LocatelloPRSBT20}
Francesco Locatello, Ben Poole, Gunnar R{\"{a}}tsch, Bernhard Sch{\"{o}}lkopf,
  Olivier Bachem, and Michael Tschannen.
\newblock Weakly-supervised disentanglement without compromises.
\newblock In {\em ICML}, pages 6348--6359, 2020.

\bibitem{RTN}
Mingsheng Long, Han Zhu, Jianmin Wang, and Michael~I. Jordan.
\newblock Unsupervised domain adaptation with residual transfer networks.
\newblock In {\em NeurIPS}, pages 136--144, 2016.

\bibitem{ParetoDA}
Fangrui Lv, Jian Liang, Kaixiong Gong, Shuang Li, Chi~Harold Liu, Han Li, Di
  Liu, and Guoren Wang.
\newblock Pareto domain adaptation.
\newblock In {\em NeurIPS}, 2021.

\bibitem{MMAN}
Xinhong Ma, Tianzhu Zhang, and Changsheng Xu.
\newblock Deep multi-modality adversarial networks for unsupervised domain
  adaptation.
\newblock {\em {IEEE} Trans. Multim.}, 21(9):2419--2431, 2019.

\bibitem{MatchDG}
Divyat Mahajan, Shruti Tople, and Amit Sharma.
\newblock Domain generalization using causal matching.
\newblock In {\em ICML}, pages 7313--7324, 2021.

\bibitem{Meinshausen2016Causal}
Meinshausen, Nicolai, Buhlmann, Peter, Peters, and Jonas.
\newblock Causal inference by using invariant prediction: identification and
  confidence intervals.
\newblock {\em Journal of the Royal Statistical Society, Series B. Statistical
  Methodology}, 2016.

\bibitem{CCSA}
Saeid Motiian, Marco Piccirilli, Donald~A. Adjeroh, and Gianfranco Doretto.
\newblock Unified deep supervised domain adaptation and generalization.
\newblock In {\em ICCV}, pages 5716--5726, 2017.

\bibitem{DICA}
Krikamol Muandet, David Balduzzi, and Bernhard Sch{\"{o}}lkopf.
\newblock Domain generalization via invariant feature representation.
\newblock In {\em ICML}, pages 10--18, 2013.

\bibitem{Sagnet}
Hyeonseob Nam, HyunJae Lee, Jongchan Park, Wonjun Yoon, and Donggeun Yoo.
\newblock Reducing domain gap via style-agnostic networks.
\newblock {\em CoRR}, abs/1910.11645, 2019.

\bibitem{SVHN}
Yuval Netzer, Tiejie Wang, Adam Coates, Alessandro Bissacco, Baolin Wu, and
  Andrew~Y. Ng.
\newblock Reading digits in natural images with unsupervised feature learning.
\newblock In {\em NeurIPS}, 2011.

\bibitem{1981Fast}
H.~J. Nussbaumer.
\newblock {\em Fast Fourier Transform and Convolution Algorithms}.
\newblock Springer-Verlag, 1981.

\bibitem{1981fourier}
A.~V. Oppenheim and J.~S. Lim.
\newblock The importance of phase in signals.
\newblock {\em Proc IEEE}, 69(5):529--541, 1981.

\bibitem{LICM}
Giambattista Parascandolo, Niki Kilbertus, Mateo Rojas{-}Carulla, and Bernhard
  Sch{\"{o}}lkopf.
\newblock Learning independent causal mechanisms.
\newblock In {\em ICML}, pages 4033--4041, 2018.

\bibitem{2000Causality}
J. Pearl.
\newblock Causality: Models, reasoning, and inference, second edition.
\newblock {\em Cambridge University Press}, 2000.

\bibitem{CORAL}
Xingchao Peng, Qinxun Bai, Xide Xia, Zijun Huang, Kate Saenko, and Bo Wang.
\newblock Moment matching for multi-source domain adaptation.
\newblock In {\em ICCV}, pages 1406--1415, 2019.

\bibitem{DGCAN}
Xingchao Peng and Kate Saenko.
\newblock Synthetic to real adaptation with generative correlation alignment
  networks.
\newblock In {\em WACV}, pages 1982--1991, 2018.

\bibitem{2000Causation}
S. Peter, G. Clark, and S. Richard.
\newblock Causation, prediction, and search.
\newblock {\em The British Journal for the Philosophy of Science}, (4):4, 2000.

\bibitem{2017Elements}
J. Peters, D. Janzing, and B Schölkopf.
\newblock Elements of causal inference - foundations and learning algorithms.
\newblock {\em MIT Press, Cambridge, MA, USA}, 2017.

\bibitem{1982fourier}
L.~N. Piotrowski and F.~W. Campbell.
\newblock A demonstration of the visual importance and flexibility of
  spatial-frequency amplitude and phase.
\newblock {\em Perception}, 11(3):337--46, 1982.

\bibitem{CSD}
Vihari Piratla, Praneeth Netrapalli, and Sunita Sarawagi.
\newblock Efficient domain generalization via common-specific low-rank
  decomposition.
\newblock In {\em ICML}, pages 7728--7738, 2020.

\bibitem{1956The}
H. Reichenbach.
\newblock {\em The Direction of Time.}
\newblock University of California Press, Berkeley, CA,, 1956.

\bibitem{Rojas-CarullaST18}
Mateo Rojas{-}Carulla, Bernhard Sch{\"{o}}lkopf, Richard~E. Turner, and Jonas
  Peters.
\newblock Invariant models for causal transfer learning.
\newblock {\em J. Mach. Learn. Res.}, 19:36:1--36:34, 2018.

\bibitem{DRO}
Shiori Sagawa, Pang~Wei Koh, Tatsunori~B. Hashimoto, and Percy Liang.
\newblock Distributionally robust neural networks for group shifts: On the
  importance of regularization for worst-case generalization.
\newblock {\em CoRR}, abs/1911.08731, 2019.

\bibitem{scm}
Bernhard Sch{\"{o}}lkopf.
\newblock Causality for machine learning.
\newblock {\em CoRR}, abs/1911.10500, 2019.

\bibitem{anti-causal}
Bernhard Sch{\"{o}}lkopf, Dominik Janzing, Jonas Peters, Eleni Sgouritsa, Kun
  Zhang, and Joris~M. Mooij.
\newblock On causal and anticausal learning.
\newblock In {\em ICML}, 2012.

\bibitem{TCRL}
Bernhard Sch{\"{o}}lkopf, Francesco Locatello, Stefan Bauer, Nan~Rosemary Ke,
  Nal Kalchbrenner, Anirudh Goyal, and Yoshua Bengio.
\newblock Towards causal representation learning.
\newblock {\em CoRR}, abs/2102.11107, 2021.

\bibitem{B2016Causal}
B Schölkopf, D. Janzing, and D. Lopezpaz.
\newblock Causal and statistical learning.
\newblock 2016.

\bibitem{Grad-CAM}
Ramprasaath~R. Selvaraju, Michael Cogswell, Abhishek Das, Ramakrishna Vedantam,
  Devi Parikh, and Dhruv Batra.
\newblock Grad-cam: Visual explanations from deep networks via gradient-based
  localization.
\newblock In {\em ICCV}, pages 618--626, 2017.

\bibitem{CrossGrad}
Shiv Shankar, Vihari Piratla, Soumen Chakrabarti, Siddhartha Chaudhuri, Preethi
  Jyothi, and Sunita Sarawagi.
\newblock Generalizing across domains via cross-gradient training.
\newblock In {\em ICLR}, 2018.

\bibitem{causality}
Ram Shanmugam.
\newblock Causality: Models, reasoning, and inference : Judea on causal and
  anticausal learning; cambridge university press, cambridge, uk, 2000, pp 384,
  {ISBN} 0-521-77362-8.
\newblock {\em Neurocomputing}, 41(1-4):189--190, 2001.

\bibitem{WSDG}
Rui Shu, Yining Chen, Abhishek Kumar, Stefano Ermon, and Ben Poole.
\newblock Weakly supervised disentanglement with guarantees.
\newblock In {\em ICLR}, 2020.

\bibitem{LPMTT}
Adarsh Subbaswamy, Peter Schulam, and Suchi Saria.
\newblock Preventing failures due to dataset shift: Learning predictive models
  that transport.
\newblock In {\em AISTATS}, pages 3118--3127, 2019.

\bibitem{robustness}
Rohan Taori, Achal Dave, Vaishaal Shankar, Nicholas Carlini, Benjamin Recht,
  and Ludwig Schmidt.
\newblock Measuring robustness to natural distribution shifts in image
  classification.
\newblock In {\em NeurIPS}, 2020.

\bibitem{ERM}
Vladimir Vapnik.
\newblock An overview of statistical learning theory.
\newblock {\em {IEEE} Trans. Neural Networks}, 10(5):988--999, 1999.

\bibitem{Office-Home}
Hemanth Venkateswara, Jose Eusebio, Shayok Chakraborty, and Sethuraman
  Panchanathan.
\newblock Deep hashing network for unsupervised domain adaptation.
\newblock In {\em CVPR}, pages 5018--5027, 2017.

\bibitem{AdvAug}
Riccardo Volpi, Hongseok Namkoong, Ozan Sener, John~C. Duchi, Vittorio Murino,
  and Silvio Savarese.
\newblock Generalizing to unseen domains via adversarial data augmentation.
\newblock In {\em NeurIPS}, pages 5339--5349, 2018.

\bibitem{EISNet}
Shujun Wang, Lequan Yu, Caizi Li, Chi{-}Wing Fu, and Pheng{-}Ann Heng.
\newblock Learning from extrinsic and intrinsic supervisions for domain
  generalization.
\newblock In {\em ECCV}, volume 12354, pages 159--176, 2020.

\bibitem{contrastive-ACE}
Yunqi Wang, Furui Liu, Zhitang Chen, Qing Lian, Shoubo Hu, Jianye Hao, and
  Yik{-}Chung Wu.
\newblock Contrastive {ACE:} domain generalization through alignment of causal
  mechanisms.
\newblock {\em CoRR}, abs/2106.00925, 2021.

\bibitem{BayesDG}
Zehao Xiao, Jiayi Shen, Xiantong Zhen, Ling Shao, and Cees Snoek.
\newblock A bit more bayesian: Domain-invariant learning with uncertainty.
\newblock In {\em ICML}, pages 11351--11361, 2021.

\bibitem{FACT}
Qinwei Xu, Ruipeng Zhang, Ya Zhang, Yanfeng Wang, and Qi Tian.
\newblock A fourier-based framework for domain generalization.
\newblock In {\em CVPR}, pages 14383--14392, 2021.

\bibitem{MIXUP}
Hongyi Zhang, Moustapha Ciss{\'{e}}, Yann~N. Dauphin, and David Lopez{-}Paz.
\newblock mixup: Beyond empirical risk minimization.
\newblock In {\em ICLR}, 2018.

\bibitem{Digits-DG}
Kaiyang Zhou, Yongxin Yang, Timothy~M. Hospedales, and Tao Xiang.
\newblock Deep domain-adversarial image generation for domain generalisation.
\newblock In {\em AAAI}, pages 130 25--13032, 2020.

\bibitem{L2A-OT}
Kaiyang Zhou, Yongxin Yang, Timothy~M. Hospedales, and Tao Xiang.
\newblock Learning to generate novel domains for domain generalization.
\newblock In {\em ECCV}, pages 561--578, 2020.

\bibitem{mini-dm}
Kaiyang Zhou, Yongxin Yang, Yu Qiao, and Tao Xiang.
\newblock Domain adaptive ensemble learning.
\newblock {\em {IEEE} Trans. Image Process.}, 30:8008--8018, 2021.

\bibitem{MixStyle}
Kaiyang Zhou, Yongxin Yang, Yu Qiao, and Tao Xiang.
\newblock Domain generalization with mixstyle.
\newblock In {\em ICLR}, 2021.

\end{thebibliography}

\clearpage
\section*{Supplementary Materials}
\appendix

\section{Potential Negative Societal Impacts}
Our work focuses on domain generalization and attempts to excavate the intrinsic causal mechanisms from a causal view, which further enhances the generalization capability of the learned model on OOD distribution.
This approach exerts a positive influence on the society and the community for saving the cost and time of data annotation, boosting the reusability of knowledge across domains, and greatly improving the generalization ability.
However, this work may also suffer from some negative impacts, which is worthy of further research and exploration. Specifically, more jobs of classification or target detection for conditions that out of the support of observed data distributions may be cancelled. 
What's more, we need to be cautious about the reliability of the system, which might be misleading when using in some conditions that are very far from the observed distributions.
\blfootnote{$*$ Corresponding author.}

\section{Implementation Details}
\subsection{Z-score Normalization}
As mentioned in the section \textit{Causal Intervention Module}, before measuring the correlation of representations before and after the intervention upon non-causal factors, we conduct Z-score normalization on the columns of representations $\Rb^o\in\mathbb{R}^{B\times N}$ and $\Rb^a$ as follows, which can convert data of different orders of magnitude into unified Z-score measurement for fair comparison:

\begin{equation}
\begin{split}
\label{Z-score}
       &\tilde{\boldsymbol{r}}_i^o = \frac{\tilde{\boldsymbol{r}}_i^o - \frac{1}{N}\sum_{i=1}^N \tilde{\boldsymbol{r}}_i^o}{\sqrt{ \frac{1}{N}\sum_{i=1}^N{(\tilde{\boldsymbol{r}}_i^o - \frac{1}{N}\sum_{i=1}^N \tilde{\boldsymbol{r}}_i^o)}^2}},\\
       &\tilde{\boldsymbol{r}}_i^a = \frac{\tilde{\boldsymbol{r}}_i^a - \frac{1}{N}\sum_{i=1}^N \tilde{\boldsymbol{r}}_i^a}{\sqrt{ \frac{1}{N}\sum_{i=1}^N{(\tilde{\boldsymbol{r}}_i^a - \frac{1}{N}\sum_{i=1}^N \tilde{\boldsymbol{r}}_i^a)}^2}},
\end{split}
\end{equation}
where $\tilde{\boldsymbol{r}}_i^o, \tilde{\boldsymbol{r}}_i^a$ denote the $i$-th column of $\Rb^o$ and $\Rb^a$ respectively.

\subsection{Gumbel-Softmax Trick}

We apply the commonly-used Gumbel-softmax~\cite{gumbel} trick to approximately sample a $k$-hot vector, where $k=\lfloor \kappa N \rfloor\in\mathbb{Z}_+$. Specifically, in Eq.~(12), let $\z=\hat{w}(\rb)\in\mathbb{R}^N$ be a probability vector, where for $j\in\{1,\ldots,N\}$, $z_j \geq 0$ and $\sum_jz_j=1$. Then we define the sampled vector $\boldsymbol{m} = \mbox{Gumbel-Softmax}(\hat{w}(\rb), \kappa N) \in \mathbb{R}^N$, where for a predefined $\tau>0,j\in\{1,\ldots,N\},l\in\{1,\ldots,k\}$,
\begin{equation}\label{eq:gumble}
\begin{split}
m_j & = \max_{l\in\{1,\ldots,k\}} \frac{\exp((\log z_j + \xi_j^l)/\tau)}{\sum_{j'=1}^N\exp((\log z_{j'} + \xi_{j'}^l)/\tau)},\\
\xi_j^l &= -\log(-\log u_j^l), \ u_j^l \sim \mbox{Uniform(0,1)}.
\end{split}
\end{equation}
We follow a common setting~\cite{chen2018learning} to let $\tau=0.5$.

\subsection{Network Structure}
\lv{
For Digits-DG, the network is the same as \cite{Digits-DG} which constructs the network with four $3\times3$ convolutional layers (each followed by ReLU and $2\times2$ max-pooling) and a softmax classification layer, denoted as ConvNet. 
As for PACS and Office-Home, we adopt ResNet \cite{resnet} pre-trained on ImageNet \cite{ImageNet} as the backbone. And the masker is implemented by a 3-layer MLP which is randomly initialized.
For fairness, the feature dimension $N$ is 256, 512, and 2048 for the experiments with backbone ConvNet, ResNet-18, and
ResNet-50 respectively following previous works.
}

\vspace{1mm}
\section{Additional Results}
\subsection{Sensitivity to Feature Dimension}
\lv{
To analyze the influence of feature dimension, we first conduct experiments for different feature dimensions $N = \{128,256,512,1024,2048\}$ with a fixed $\kappa = 60\%$ on PACS to explore the influence of feature dimension on model performance, as Table~\ref{tab:dim_model} shows. It is clear that the performance of model remains relatively stable when the feature dimension varies, which demonstrates the stability of our method.
And then we conduct experiments for fixed feature dimensions $N = 512$ and $N = 2048$ with different $\kappa = \{50\%, 60\%, 70\%, 80\%, 90\%\}$ on PACS to investigate the influence of feature dimension on the choice of $\kappa$, the results are shown in Table~\ref{tab:dim_k}. We can see that $\kappa$ is shown not sensitive to feature dimension, $\kappa = 60\%$ is a stable choice for different feature dimensions.

}

\begin{table}[!htbp]\small
  \setlength{\abovecaptionskip}{0.cm}
  \setlength{\belowcaptionskip}{0.cm}
  \centering
  \caption{Model performance sensitivity to feature dimension (ResNet-18). 
  }
    \resizebox{\columnwidth}{!}{\begin{tabular}{l|cccc|c}
    \toprule
    $N$ & Art & Cartoon & Photo & Sketch & Avg. \\
    \midrule
    128  & 85.9 & 80.1 & 95.9 & 82.6 & 86.12 \\
    256  & 85.8 & 80.2 & \textbf{96.4} & 82.6 & 86.25 \\
    512  & \textbf{86.1} & \textbf{80.6} & 95.9 & \textbf{82.7} & \textbf{86.32} \\
    1024  & 85.9 & 80.4 & 95.7 & \textbf{82.7} & 86.17 \\
    2048  & 86.0 & 80.2 & 96.3 & 82.4 & 86.23 \\
    \bottomrule
    \end{tabular}}
  \label{tab:dim_model}
\end{table}

\begin{table}[!htbp]\small
  \setlength{\abovecaptionskip}{0.cm}
  \setlength{\belowcaptionskip}{0.cm}
  \centering
  \caption{$\kappa$ selection sensitivity to feature dimension (ResNet-18). 
  }
    \resizebox{\columnwidth}{!}{
    \begin{tabular}{l|cccc|c}
    \toprule
    $\kappa$ & Art & Cartoon & Photo & Sketch & Avg.\\
    \midrule
    \midrule
    \multicolumn{6}{c}{\textit{$N = 512$}}\\
    \midrule
    50\%  & 85.5 & 79.5 & \textbf{96.2} & 82.0 & 85.80\\
    60\%  & \textbf{86.1} & 80.6 & 95.9 & \textbf{82.7} & \textbf{86.32}\\
    70\%  & 85.7 &	\textbf{80.8} &	95.8 &	82.1 &	86.13\\
    80\%  & 85.4 &	79.6 &	95.6 &	81.8 &	85.60\\
    90\%  & 85.3 &	79.6 &	95.6 &	81.4 &	85.48\\
    \midrule
    \midrule
    \multicolumn{6}{c}{\textit{$N = 2048$}}\\
    \midrule
     50\%  & 85.7 & 79.8 & 96.2 & \textbf{82.6} & 86.08\\
    60\%  & \textbf{86.0} & \textbf{80.2} & \textbf{96.3} & 82.4 & \textbf{86.23}\\
    70\%  & 85.6 & 80.1 & 96.0 & 81.7 & 85.85\\
    80\%  & 84.7 & 78.9 & 95.9 & 82.1 & 85.40\\
    90\%  & 84.6 & 79.7 & 95.8 & 81.3 & 85.35\\

    \bottomrule
    \end{tabular}}
  \label{tab:dim_k}
\end{table}


\subsection{Sampling Strategies for Amplitude Mixing}
\lv{
In the causal intervention module, we mix amplitude spectrum of a specific image and another image sampled randomly from arbitrary source domains. Nevertheless, we can also restrict the sample pair to be taken from the same domain (intra-domain) or different domains (inter-domain).
To explore the effect of different sampling strategies, we conduct experiments by using only intra-domain or inter-domain strategy on PACS with ResNet-18, as Table~\ref{tab:sample_strategy} shows.
As we can see, CIRL is shown not sensitive to the sampling strategies. Whether intra- or inter-domain sampling strategy brings a good performance, and using a fully random strategy works best, perhaps because more causal interventions are conducted which leads to extracting better causal features.
}

\begin{table}[!htbp]\small
  \setlength{\abovecaptionskip}{0.cm}
  \setlength{\belowcaptionskip}{0.cm}
  \centering
  \caption{Impact of sampling strategies for amplitude mixing. 
  }
    \resizebox{\columnwidth}{!}{\begin{tabular}{l|cccc|c}
    \toprule
    Sampling Strategy & Art & Cartoon & Photo & Sketch & Avg. \\
    \midrule
    intra-domain  & 84.92 & 80.76& 96.23& 82.77& 86.17 \\
    inter-domain  & 85.45 & 79.93& 96.53& 82.14& 86.01 \\
    random & 86.08 & 80.59 & 95.93 & 82.67 & 86.32 \\
    \bottomrule
    \end{tabular}}
  \label{tab:sample_strategy}
\end{table}


\subsection{Experimental Results on mini-DomainNet}
\lv{
Note that the benchmarks used for experiments in the main body are all relatively small-scale datasets, which may be fairly saturated in performance so that the improvement of our CIRL seems to be incremental.
Thus, we additionally carry out experiments on a large-scale dataset
mini-DomainNet following~\cite{mini-dm} which has
140K images with 126 classes. The results are shown in Table~\ref{tab:mini-dm}, we can see that CIRL outperforms the SOTA method by a large margin of 3.26\%, which validates our superiority.

\begin{table}[!htbp]\small
  \setlength{\abovecaptionskip}{0.cm}
  \setlength{\belowcaptionskip}{0.cm}
  \centering
  \caption{Leave-one-domain-out results on mini-DomainNet with ResNet-18. 
  }
    \resizebox{\columnwidth}{!}{\begin{tabular}{l|cccc|c}
    \toprule
    Method & Clipart & Painting & Real & Sketch & Avg. \\
    \midrule
    ERM~\cite{ERM} &65.5 & 57.1 & 62.3 & 57.1 & 60.50\\
    DRO~\cite{DRO} &64.8 & 57.4 & 61.5 & 56.9 & 60.15\\
    Mixup~\cite{MIXUP} &\underline{67.1} & \underline{59.1} & 64.3& \underline{59.2} & 62.42\\
    MLDG~\cite{MLDG} &65.7 &57.0 & 63.7 & 58.1 & 61.12\\
    CORAL~\cite{CORAL} & 66.5 & 59.5 & \underline{66.0} & 59.5 & \underline{62.87}\\
    MMD~\cite{MMD} &65.0 & 58.0 &63.8 & 58.4& 61.30\\
    MTL~\cite{MTL} & 65.3 & 59.0& 65.6 & 58.5&62.10\\
    SagNet~\cite{Sagnet} &65.0 & 58.1 & 64.2 & 58.1 & 61.35\\
    CIRL (\textit{ours}) & \textbf{70.2} & \textbf{62.9}& \textbf{67.8}& \textbf{63.6}& \textbf{66.13}\\
    \bottomrule
    \end{tabular}}
  \label{tab:mini-dm}
\end{table}
}

\begin{figure*}[!h]
  \centering
  \includegraphics[width=1.0\linewidth]{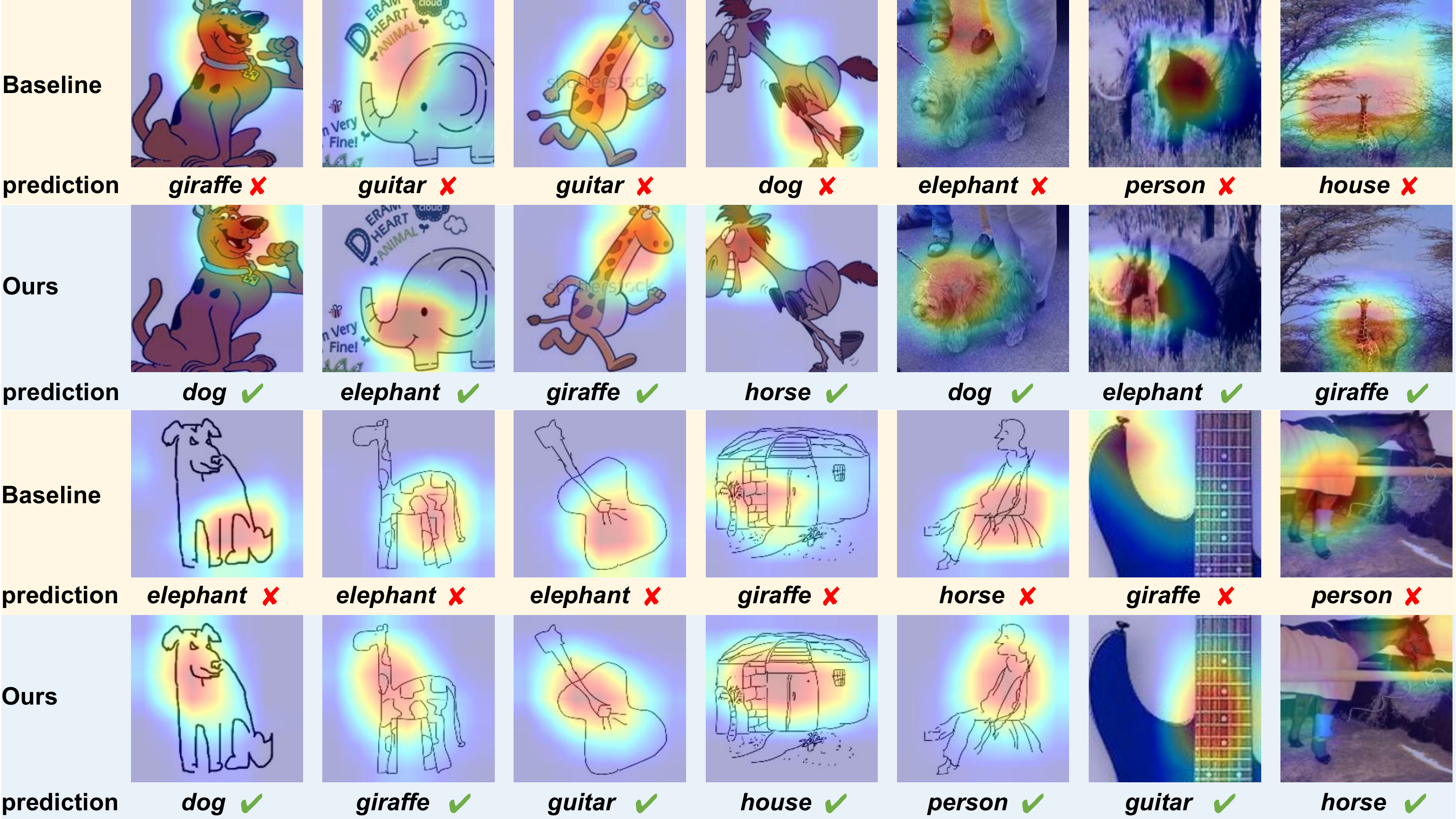}
   \caption{More examples of the visualization explanation on PACS dataset, with Cartoon, Photo, Sketch as the unseen target domain respectively for baseline (i.e., DeepAll) and CIRL methods.}
   \label{fig:attention_sup}
\end{figure*}

\begin{table*}[!htbp]
\caption{Leave-one-domain-out results on Digits-DG. 
}
  \centering
    \begin{tabular}{l|cccc|c}
    \toprule
    Methods & MNIST & MNIST-M & SVHN & SYN & Avg. \\
    \midrule
    Jigen~\cite{Jigen} & 96.5& 61.4 & 63.7 & 74.0 & 73.9 \\
     L2A-OT~\cite{L2A-OT} & 96.7 & 63.9 &68.6& 83.2 & 78.1 \\
     \midrule
    DeepAll~\cite{Digits-DG} &95.8$\pm$0.3 & 58.8$\pm$0.5 & 61.7$\pm$0.5 & 78.6$\pm$0.6 & 73.7\\
    CCSA~\cite{CCSA} & 95.2$\pm$0.2 & 58.2$\pm$0.6 & 65.5$\pm$0.2 & 79.1$\pm$0.8 & 74.5 \\
    MMD-AAE~\cite{MMD-AAE} & 96.5$\pm$0.1 & 58.4$\pm$0.1 & 65.0$\pm$0.1 & 78.4$\pm$0.2 & 74.6 \\
    CrossGrad~\cite{CrossGrad} &96.7$\pm$0.1 & 61.1$\pm$0.5 & 65.3$\pm$0.5 & 80.2$\pm$0.2 & 75.8 \\
    DDAIG~\cite{Digits-DG} & 96.6$\pm$0.2 & 64.1$\pm$0.4 & 68.6$\pm$0.6 & 81.0$\pm$0.5 & 77.6 \\
    FACT \cite{FACT} & \textbf{97.9$\pm$0.2} & 65.6$\pm$0.4 & 72.4$\pm$0.7 & \textbf{90.3$\pm$0.1} & 81.5  \\
    \midrule
    CIRL (\textit{ours}) & 96.08$\pm$0.2 &	\textbf{69.87 $\pm$ 0.5}&	\textbf{76.17 $\pm$ 0.4}&	87.68$\pm$0.3 &	\textbf{82.5} \\
    \bottomrule
    \end{tabular}
  \label{digitsdg-full}
\end{table*}

\begin{table*}[!htbp]
  \centering
  \caption{Leave-one-domain-out results on PACS with ResNet-18. 
  }
    \begin{tabular}{l|cccc|c}
    \toprule
    Methods & Art & Cartoon & Photo & Sketch & Avg. \\
    \midrule
    JiGen~\cite{Jigen} & 79.42 & 75.25 & 96.03 & 71.35 & 80.51 \\
    L2A-OT~\cite{L2A-OT} & 83.30 & 78.20 & 96.20 & 73.60 & 82.80 \\
    RSC~\cite{RSC} & 83.43 & 80.31 & 95.99 & 80.85 & 85.15 \\
    \midrule
    DeepAll\cite{Digits-DG} &  77.63$\pm$0.84 & 76.77$\pm$0.33 & 95.85$\pm$0.20 & 69.50$\pm$1.26 & 79.94 \\
    MetaReg~\cite{MetaReg} & 83.70$\pm$0.19 & 77.20$\pm$0.31 & 95.50$\pm$0.24 & 70.30$\pm$0.28 & 81.70 \\
    DDAIG~\cite{Digits-DG} & 84.20$\pm$0.30 & 78.10$\pm$0.60 & 95.30$\pm$0.40 & 74.70$\pm$0.80 & 83.10 \\
    CSD~\cite{CSD} & 78.90$\pm$1.10 & 75.80$\pm$1.00 & 94.10$\pm$0.20 & 76.70$\pm$1.20 & 81.40 \\
    MASF~\cite{MASF} & 80.29$\pm$0.18 & 77.17$\pm$0.08 & 94.99$\pm$0.09 & 71.69$\pm$0.22&  81.04 \\
    EISNet~\cite{EISNet} & 81.89$\pm$0.88 & 76.44$\pm$0.31 & 95.93$\pm$0.06 & 74.33$\pm$1.37 & 82.15 \\
    MatchDG \cite{MatchDG} & 81.32$\pm$0.38 & \textbf{80.70 $\pm$ 0.54} & 96.53$\pm$0.05 & 79.72 $\pm$ 1.01 & 84.56 \\
    FACT \cite{FACT}& 85.90$\pm$0.27 & 79.35$\pm$0.03 & \textbf{96.61$\pm$0.17} &80.89$\pm$0.26 & 85.69\\
    \midrule
    CIRL (\textit{ours})  & \textbf{86.08$\pm$0.32} & 80.59$\pm$0.19 & 95.93$\pm$0.03 & \textbf{82.67$\pm$0.47} & \textbf{86.32} \\
    \bottomrule
    \end{tabular}
  \label{pacs-res18-full}
\end{table*}

\begin{table*}[!htbp]
  \centering
  \caption{Leave-one-domain-out results on PACS with ResNet-50. 
  }
    \begin{tabular}{l|cccc|c}
    \toprule
    Methods & Art & Cartoon & Photo & Sketch & Avg. \\
    \midrule
    RSC~\cite{RSC} & 87.89 & 82.16 & 97.92 & 83.35 & 87.83 \\
    \midrule
    DeepAll\cite{Digits-DG} & 84.94$\pm$0.66 & 76.98$\pm$1.13 & 97.64$\pm$0.10 & 76.75$\pm$0.41 & 84.08 \\
    MetaReg~\cite{MetaReg} & 87.20$\pm$0.13 & 79.20$\pm$0.27 & 97.60$\pm$0.31 & 70.30$\pm$0.18 & 83.60 \\
    MASF~\cite{MASF}  & 82.89$\pm$0.16 & 80.49$\pm$0.21 & 95.01$\pm$0.10 & 72.29$\pm$0.15 & 82.67 \\
    EISNet~\cite{EISNet} & 86.64$\pm$1.41 & 81.53$\pm$0.64 & 97.11$\pm$0.40 & 78.07$\pm$1.43 & 85.84 \\
    MatchDG \cite{MatchDG} & 85.61$\pm$0.81 & 82.12$\pm$0.69 & \textbf{97.94$\pm$0.27} & 78.76$\pm$1.13 & 86.11 \\
    FACT \cite{FACT} &\textbf{90.89$\pm$0.19} & 83.65$\pm$0.12 & 97.78$\pm$0.05 & 86.17$\pm$0.14 & 89.62 \\
    \midrule
    CIRL (\textit{ours}) & 90.67$\pm$0.21 & \textbf{84.30$\pm$0.17} & 97.84$\pm$0.08 & \textbf{87.68$\pm$0.40} & \textbf{90.12} \\
    \bottomrule
    \end{tabular}
  \label{pacs-50-full}
\end{table*}

\begin{table*}[!htbp]
  \centering
  \caption{Leave-one-domain-out results on Office-Home with ResNet-18.
}
    \begin{tabular}{l|cccc|c}
    \toprule
    Methods & Art & Clipart & Product & Real & Avg. \\
    \midrule
     Jigen~\cite{Jigen} & 53.04 & 47.51 & 71.47 & 72.79 & 61.20 \\
    RSC~\cite{RSC}   & 58.42 & 47.90 & 71.63 & 74.54 & 63.12 \\
    L2A-OT~\cite{L2A-OT} & 60.60 & 50.10 & 74.80 & \textbf{77.00} & 65.60 \\
    \midrule
    DeepAll \cite{Digits-DG} & 57.88$\pm$0.20 & 52.72$\pm$0.50 & 73.50$\pm$0.30 & 74.80$\pm$0.10 & 64.72 \\
    CCSA~\cite{CCSA}  & 59.90$\pm$0.30 & 49.90$\pm$0.40 & 74.10$\pm$0.20 & 75.70$\pm$0.20 & 64.90 \\
    MMD-AAE~\cite{MMD-AAE} & 56.50$\pm$0.40 & 47.30$\pm$0.30 & 72.10$\pm$0.30 & 74.80$\pm$0.20 & 62.70 \\
    CrossGrad~\cite{CrossGrad} & 58.40$\pm$0.70 & 49.40$\pm$0.40 & 73.90$\pm$0.20 & 75.80$\pm$0.10 & 64.40 \\
    DDAIG~\cite{Digits-DG} & 59.20$\pm$0.10 & 52.30$\pm$0.30 & 74.60$\pm$0.30 & 76.00$\pm$0.10 & 65.50 \\
    FACT \cite{FACT}& 60.34$\pm$0.11 & 54.85$\pm$0.37 & 74.48$\pm$0.13 & 76.55$\pm$0.10 & 66.56 \\
    \midrule
     CIRL (\textit{ours}) & \textbf{61.48$\pm$0.17} & \textbf{55.28$\pm$0.29} & \textbf{75.06$\pm$0.24} & 76.64$\pm$0.09 & \textbf{67.12} \\
    \bottomrule
    \end{tabular}
  \label{officehome-full}
\end{table*}

\subsection{More Examples for Visual Explanation}
We present more examples of attention maps of the last convolutional layer for baseline (i.e., DeepAll) and CIRL methods in Fig. \ref{fig:attention_sup}, utilizing the visualization technique in \cite{Grad-CAM}.
As we can see, in all the tasks, the representations learned by CIRL precisely capture the category-related information of different objects that contributes to the classification, while the representations learned by baseline method contain many noisy information such as background that leads to wrong predictions.

\subsection{Experimental Results with Error Bars}
For the sake of objective, we run all the experiments multiple times with random seed. We report the average results in the main body of paper for elegant, and show the complete results with error bars in the form of mean$\pm$std below (Table. \ref{digitsdg-full},\ref{pacs-res18-full},\ref{pacs-50-full},\ref{officehome-full}). 
\vspace{4mm}


\end{document}